%% file: main.tex
\documentclass[10pt,twocolumn,letterpaper]{article}

\usepackage{cvpr}              
\input{preamble}
\definecolor{cvprblue}{rgb}{0.21,0.49,0.74}
\usepackage[pagebackref,breaklinks,colorlinks,allcolors=cvprblue]{hyperref}
\usepackage{booktabs}
\usepackage{multirow}
\usepackage{colortbl}
\usepackage[table]{xcolor}
\usepackage{siunitx}
\usepackage{caption}
\usepackage{array}
\usepackage{cuted} 
\usepackage{capt-of} 
\usepackage{algorithm}
\usepackage{algpseudocode}
\usepackage{float}
\usepackage[resetlabels]{multibib}
\newcites{supp}{References}

\def\paperID{*****} 
\def\confName{CVPR}
\def\confYear{2026}

\title{MMFace-DiT: A Dual-Stream Diffusion Transformer for High-Fidelity Multimodal Face Generation}

\author{Bharath Krishnamurthy \qquad Ajita Rattani\\
University of North Texas\\
Denton, TX, USA\\
{\tt\small bharathkrishnamurthy@my.unt.edu, ajita.rattani@unt.edu}
}

\begin{document}
\maketitle

\begin{strip}
    \centering
    \includegraphics[width=0.90\textwidth]{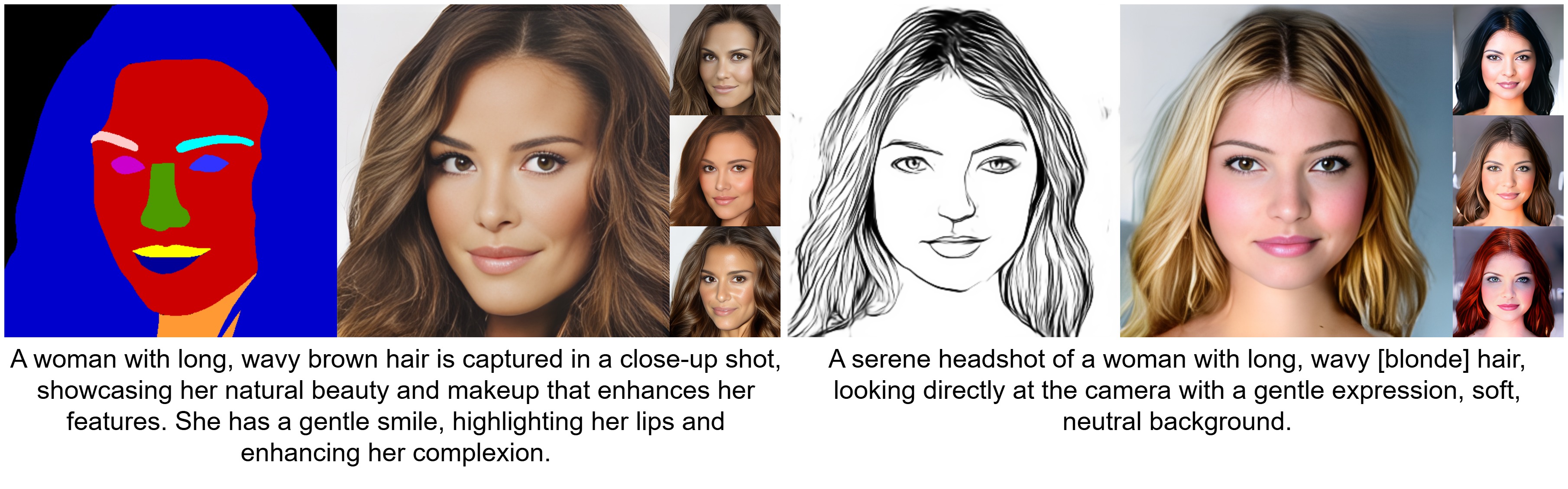}
    \captionof{figure}{\textbf{High-Fidelity Face Synthesis.} MMFace-DiT synthesizes photorealistic portraits from multi-modal inputs. \textbf{Left:} Given a semantic mask and text prompt, our model generates a face with diverse identity variations across multiple VAEs. \textbf{Right:} Guided by a sketch, it performs precise attribute-guided generation for numerous hair colors, namely, yellow, black, brown, and red. This demonstrates our model's ability to seamlessly fuse spatial and semantic guidance.}
    \label{fig:teaser}
\end{strip}

\input{sec/0_abstract}

\input{sec/1_intro}
\input{sec/2_related_work}
\input{sec/3_methodology}
\input{sec/4_experiments}
\input{sec/5_Ablation}
\input{sec/6_conclusion}
{
    \small
    \bibliographystyle{ieeenat_fullname}
    \bibliography{main}
}

\input{sec/X_suppl}

\end{document}

%% file: sec/0_abstract.tex

\begin{abstract}
Recent multimodal face generation models address the spatial control limitations of text-to-image diffusion models by augmenting text-based conditioning with spatial priors such as segmentation masks, sketches, or edge maps. This multimodal fusion enables controllable synthesis aligned with both high-level semantic intent and low-level structural layout. However, most existing approaches typically extend pre-trained text-to-image pipelines by appending auxiliary control modules or stitching together separate uni-modal networks. These ad hoc designs inherit architectural constraints, duplicate parameters, and often fail under conflicting modalities or mismatched latent spaces, limiting their ability to perform synergistic fusion across semantic and spatial domains. We introduce MMFace-DiT, a unified dual-stream diffusion transformer engineered for synergistic multimodal face synthesis. Its core novelty lies in a dual-stream transformer block that processes spatial (mask/sketch) and semantic (text) tokens in parallel, deeply fusing them through a shared Rotary Position-Embedded (RoPE) Attention mechanism. This design prevents modal dominance and ensures strong adherence to both text and structural priors to achieve unprecedented spatial–semantic consistency for controllable face generation. Furthermore, a novel Modality Embedder enables a single cohesive model to dynamically adapt to varying spatial conditions without retraining. MMFace-DiT achieves a $40\%$ improvement in visual fidelity and prompt alignment over six state-of-the-art multimodal face generation models, establishing a flexible new paradigm for end-to-end controllable generative modeling. The code and dataset are available on our project page:~\href{https://vcbsl.github.io/MMFace-DiT/}{vcbsl/MMFace-DiT}
\end{abstract}

\vspace{-1em}

%% file: sec/1_intro.tex
\section{Introduction}
\label{sec:intro}
The advent of diffusion models has revolutionized generative AI, driving major advances in text-to-image (T2I) synthesis. This progress began with a paradigm shift away from traditional GAN models~\cite{zhang2017stackgan, zhang2018stackgan++, zhu2019dm, tao2022df, ma2024prompt} towards powerful U-Net-based diffusion architectures, as seen in foundational models such as Stable Diffusion~\cite{rombach2022high, podell2023sdxl}, followed by more scalable and powerful Diffusion Transformers (DiTs)~\cite{DiT2023, esser2024scaling, wu2025qwen, cai2025hidream, li2024hunyuan}. Despite their impressive generative quality and scalability, current diffusion models still lack mechanisms for precise spatial control—\textbf{limiting} their effectiveness in structured or creative synthesis tasks, such as controllable face generation, that demand explicit spatial–semantic alignment. 

In contrast, the domain of \textit{multimodal controllable face generation} seeks to bridge this gap by integrating semantic, spatial, and structural conditioning across diverse modalities. However, existing approaches remain constrained by both design and data limitations. GAN-based controllable face generation models~\cite{xia2021tedigan, kim2024diffusion} suffer from entangled latent spaces, hindering the representation of fine-grained attributes such as earrings, hats, or accessories~\cite{karras2019style}. Conditioning adapters like ControlNet~\cite{zhang2023adding} retrofit pre-trained diffusion backbones for spatial conditioning, yet frozen parameters limit deep semantic–spatial fusion. Meanwhile, \textit{inference-time} compositional frameworks~\cite{nair2023unite, huang2023collaborative} attempt to combine uni-modal generators, but often fail under conflicting modalities (e.g., a “long hair” prompt applied to a male mask) and enforce rigid architectural constraints such as matched latent dimensionality. A recurring \textbf{limitation} across these paradigms is the trade-off between spatial fidelity and semantic consistency, where improving structural accuracy compromises textual or attribute adherence. These challenges are compounded by the scarcity of large-scale, semantically annotated face datasets: CelebA-HQ~\cite{xia2021tedigan} captions are semantically shallow, while FFHQ~\cite{karras2019style} lacks annotations altogether, impeding progress in multimodal face generation.


To address these interconnected challenges, we propose the \textbf{Dual-Stream Multi-Modal Diffusion Transformer (MMFace-DiT)}, a unified, end-to-end model that establishes a new paradigm for native multi-modal integration. Unlike auxiliary add-ons or compositional approaches, our model jointly processes and fuses semantic (text) and spatial (masks, sketches) conditions (see Figure~\ref{fig:teaser}). To solve compromised prompt adherence, its \textbf{dual-stream design} treats these conditions as co-equals, processing them in parallel and deeply fusing them at every block via a shared \textbf{RoPE Attention} mechanism to improve cross-modal alignment. Finally, we overcome the \textbf{dataset bottleneck} through a robust annotation pipeline built on the InternVL3~\cite{zhu2025internvl3} Visual Language Model (VLM). Leveraging a multi-prompt strategy with rigorous post-processing, we curate and release a large-scale, semantically rich face dataset to aid research in multimodal face generation. 

The core \textbf{contributions} of our work are summarized as follows:
\begin{enumerate}
\item \textbf{Novel Multimodal Architecture.} A unified transformer that jointly processes spatial and semantic modalities without separate models or inference-time composition.
\item \textbf{Cross-Modal Fusion.} Shared RoPE Attention that aligns and fuses text and image streams at every block for superior prompt adherence.
\item \textbf{Dynamic Modality Embedding.} A novel Modality Embedder that allows a single model to dynamically interpret different spatial conditions (e.g., masks or sketches) without retraining.
\item \textbf{Richly Annotated Face Dataset.} A large-scale, semantically rich extension of FFHQ and CelebA-HQ, annotated via a VLM-based multi-prompt pipeline to aid multimodal face generation research.
\end{enumerate}

\begin{figure*}[ht]
    \centering
    \includegraphics[width=0.95\textwidth]{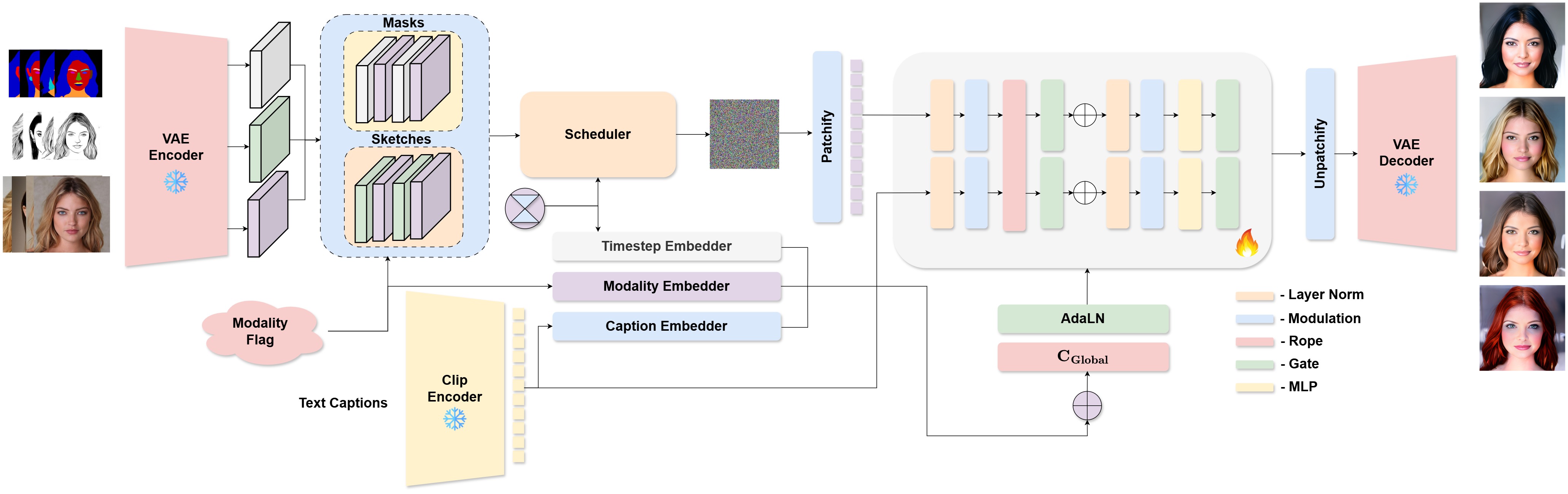}
    \caption{\textbf{Overview of the MMFace-DiT Generation Pipeline.} Our model operates in a VAE's latent space. During a forward pass, a noisy latent image is converted into a sequence of image tokens. Concurrently, a text prompt is encoded into text tokens by a CLIP encoder, which also produces pooled embeddings for global conditioning. A global conditioning vector, $C_{\text{global}}$, is formed by combining embeddings from the timestep, the text caption, and our novel Modality Embedder, which processes a flag indicating the spatial condition type. The image tokens, text tokens, and $C_{\text{global}}$ are then processed by our core transformer block, which predicts either the noise $\epsilon$ (DDPM) or the velocity $v$ (RFM). The final image is produced by unpatchifying the output tokens and decoding them using the VAE.}
    \label{fig:Overview}
\end{figure*}

%% file: sec/2_related_work.tex
\section{Related Work}
\label{sec:related_work}

\subsection{The Rise of DiTs for Image Generation}

Diffusion Probabilistic Models (DPMs) have become the leading paradigm for high-quality image generation, evolving from early work such as DDPM \cite{ho2020denoising, nichol2021glide, saharia2022photorealistic} to more efficient Latent Diffusion Models (LDMs)~\cite{rombach2022high, podell2023sdxl, wu2025qwen, cai2025hidream}, which operate in a compressed latent space. For years, the U-Net architecture was the de facto standard for the denoising network. However, the introduction of DiT~\cite{DiT2023} marked a pivotal moment, demonstrating that a transformer-based backbone could not only replace U-Net but also offer superior scalability and performance. This architectural shift has unlocked new levels of quality and coherence in T2I synthesis, powering the latest generation of state-of-the-art models such as PixArt-$\alpha$ \cite{chen2023pixart} and Stable Diffusion 3~\cite{esser2024scaling}. Building upon this foundation, we \textit{specialize} the DiT backbone for multi-modal face generation, enabling joint semantic–spatial reasoning within a unified generative framework.

\subsection{Architectures for Multi-Modal Control}
While modern DiTs excel at text-to-image synthesis, achieving precise spatial control requires conditioning on additional inputs such as masks or sketches~\cite{xia2021tedigan, xia2021towards, zhang2023adding, du2023pixelface+, meng2024mm2latent}. Various strategies have been proposed, each with significant architectural trade-offs.

\vspace{-1em}
\paragraph{GAN-Based Methods.} Methods like TediGAN~\cite{xia2021tedigan} and MM2Latent~\cite{meng2024mm2latent} rely on StyleGAN latent manipulation, which suffers from entangled representations failing to represent fine-grained facial attributes and accessories such as earrings, necklaces, or hats, limiting photorealism. Hybrid approaches like Diffusion-Driven GAN Inversion (DDGI)~\cite{kim2024diffusion} inherit similar limitations.

\vspace{-1em}

\paragraph{Conditioning Adapters.} ControlNet~\cite{zhang2023adding} introduces spatial control by attaching trainable auxiliary modules to large, pre-trained T2I diffusion models. While this retrofit enhances spatial guidance, it remains constrained by the frozen backbone, preventing deep, bidirectional fusion and limiting the model’s ability to co-adapt semantic and spatial features during generation.

\vspace{-1em}

\paragraph{Compositional Frameworks.} Another line of work focuses on the inference-time composition of multiple pre-trained, single-purpose models~\cite{nair2023unite, huang2023collaborative}. These methods are often bottlenecked by the weakest constituent model and impose rigid constraints, such as requiring identical latent space dimensions~\cite{nair2023unite}, which can fail when modalities present conflicting information.





%% file: sec/3_methodology.tex
\section{Proposed Methodology}
\label{sec:methodology}

\begin{figure}[ht]
    \centering
    \includegraphics[width=0.40\textwidth]{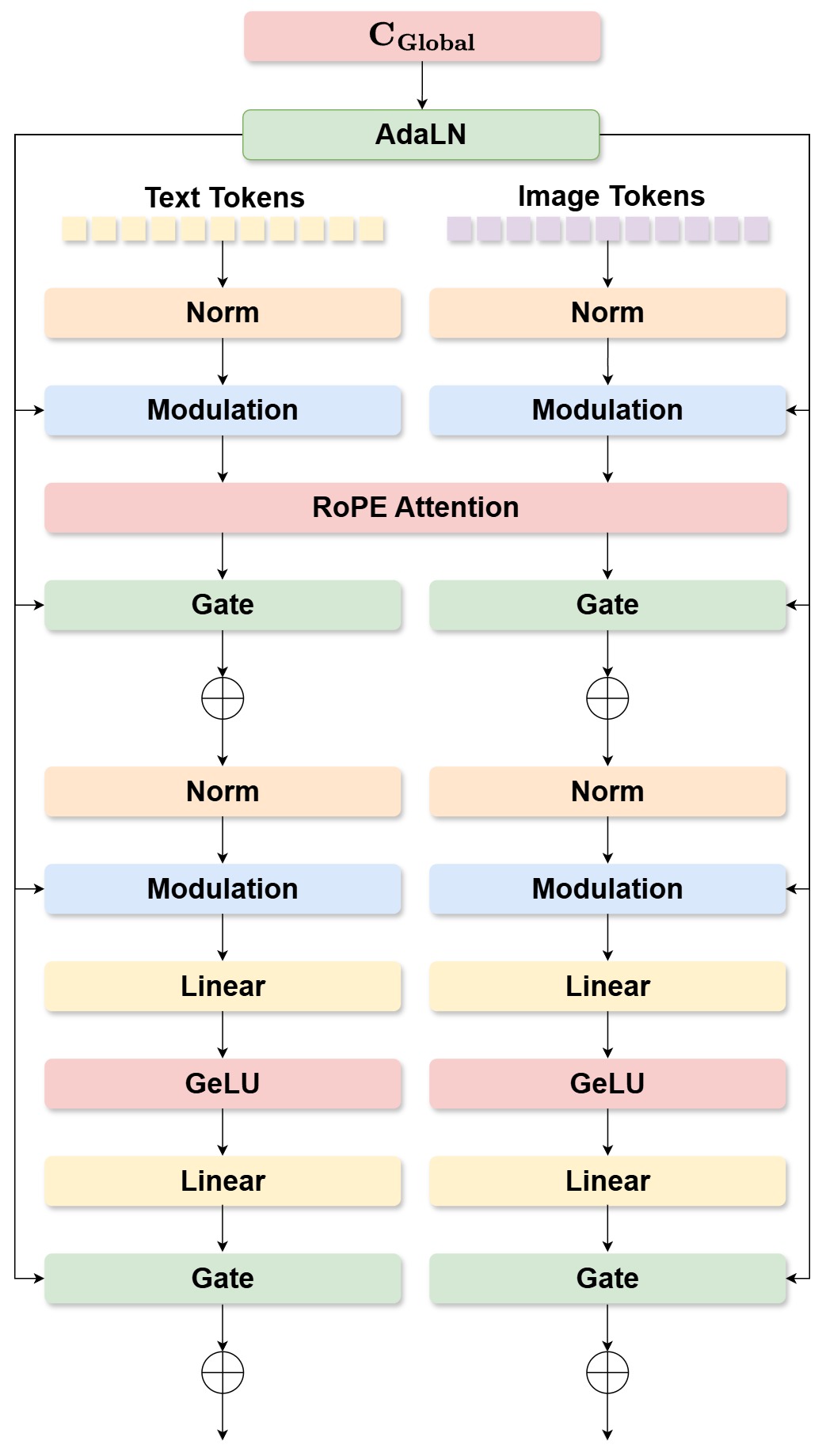}
    \caption{\textbf{Architecture of the MMFace-DiT Block.} The block processes image and text tokens in parallel, modulated by a global conditioning vector ($C_{\text{global}}$) via AdaLN. A shared RoPE Attention layer acts as the central fusion mechanism for deep cross-modal interaction. Following attention and MLP operations, each stream is processed and controlled by a gated residual connection.}
    \label{fig:block}
\end{figure}

We introduce \textbf{MMFace-DiT}, a unified, end-to-end diffusion transformer that natively processes textual descriptions alongside dynamically selected spatial conditions (masks or sketches) for high-fidelity, controllable face synthesis. Depicted in Figure~\ref{fig:Overview}, our approach operates in a VAE's latent space, guided by a unified conditioning signal. Its core novelty is a transformer backbone with co-equal spatial and semantic streams, which are deeply fused at every layer via shared attention. This design directly addresses the critical challenge of maintaining high fidelity across all input modalities simultaneously.

\subsection{Data and Architectural Preliminaries}
Our model is built upon a latent diffusion framework, leveraging robust pre-trained encoders for feature extraction and a novel data annotation pipeline, discussed in detail below:

\paragraph{VLM-Powered Data Enrichment.} To address the lack of semantically rich prompts for high-fidelity face generation, we construct a large-scale caption dataset for FFHQ and CelebA-HQ using the \textbf{InternVL3} VLM~\cite{zhu2025internvl3}. Our multi-prompt strategy—ten engineered prompts per image—captures both natural descriptions and structured demographic cues. Generated outputs undergo a two-stage refinement: a rule-based filter removes artifacts, and \textbf{Qwen3}~\cite{yang2025qwen3} Language model performs post-processing to reduce VLM hallucinations and improve factual consistency. Capped at 77 tokens per sample, the pipeline yields 1M high-quality captions (10 per image for 100K images), publicly released to support future multi-modal research. Additional implementation details are provided in the Supplementary Material.

\vspace{-1em}

\paragraph{Latent Space Diffusion.} We operate in the compressed latent space of a powerful VAE to ensure computational tractability without sacrificing visual quality. An input face image $x \in \mathbb{R}^{H \times W \times 3}$ is mapped to a latent representation $z = \mathcal{E}_{\text{vae}}(x)$, where $z \in \mathbb{R}^{h \times w \times c}$. The channel dimension $c$ depends on the VAE architecture: we explore both the Stable Diffusion VAE with $c=4$ channels and the FLUX VAE with $c=16$ channels in our ablation studies. Similarly, any spatial conditioning input $c_{\text{sp}} \in \{c_{\text{mask}}, c_{\text{sketch}}\}$ is encoded into a corresponding conditioning latent $z_c = \mathcal{E}_{\text{vae}}(c_{\text{sp}})$. The concatenation of image and conditioning latents results in an input tensor with $2c$ channels (e.g., 8 channels for SD VAE, 32 channels for FLUX VAE). All subsequent diffusion and denoising operations occur in this unified compact latent space.


\vspace{-1em}

\paragraph{Multi-Faceted Textual Embeddings.} A text prompt $p$, sampled from our VLM-generated annotations, is encoded by a pre-trained CLIP text encoder ($\mathcal{E}_{\text{text}}$), producing two complementary representations: (i) a pooled embedding $c_{\text{pooled}}$, derived from the final hidden state of the [CLS] token to capture global semantics, and (ii) a sequence of token embeddings $c_{\text{seq}}$, extracted from the penultimate layer to retain fine-grained contextual information.

\vspace{-1em}

\paragraph{Input Tokenization.} The forward pass begins by creating token sequences. The noisy image latent $z_t$ and the spatial condition latent $z_c$ are concatenated channel-wise. A patch embedding layer projects this combined tensor into a sequence of flattened \textit{image tokens} $T_i \in \mathbb{R}^{N \times D}$, where $N$ is the number of patches and $D$ is the hidden dimension. Concurrently, the CLIP sequence embeddings $c_{\text{seq}}$ are linearly projected to form the \textit{text tokens} $T_t \in \mathbb{R}^{L \times D}$, where $L$ is the sequence length.

\subsection{Unified Conditioning and Dynamic Modality Adaptation}

A key innovation of our model is its ability to adapt to different spatial modalities within a single forward pass, driven by a unified global conditioning signal. We formulate a global conditioning vector, $C_{\text{global}}$, that consolidates all non-tokenized information:
\begin{equation}
C_{\text{global}} = E_{\text{time}}(t) + E_{\text{caption}}(c_{\text{pooled}}) + E_{\text{modality}}(m)
\label{eq:c_global}
\end{equation}
Here, $E_{\text{time}}$ is a sinusoidal timestep embedder, and $E_{\text{caption}}$ is an MLP projecting the pooled CLIP embedding. The critical novel component is our \textbf{Modality Embedder}, $E_{\text{modality}}$. This is a lightweight yet highly effective embedding layer that maps a discrete modality flag $m$ (e.g., $0$ for mask, $1$ for sketch) to a dense vector in $\mathbb{R}^D$. Critically, this allows a \textbf{single set of model weights} to dynamically process different spatial conditions without retraining, unlike prior works that require separate models per modality. By injecting this modality-specific signal directly into the global context, we empower our architecture to reconfigure its processing based on the input type.

\begin{figure*}[ht]
\centering
\includegraphics[width=0.90\textwidth]{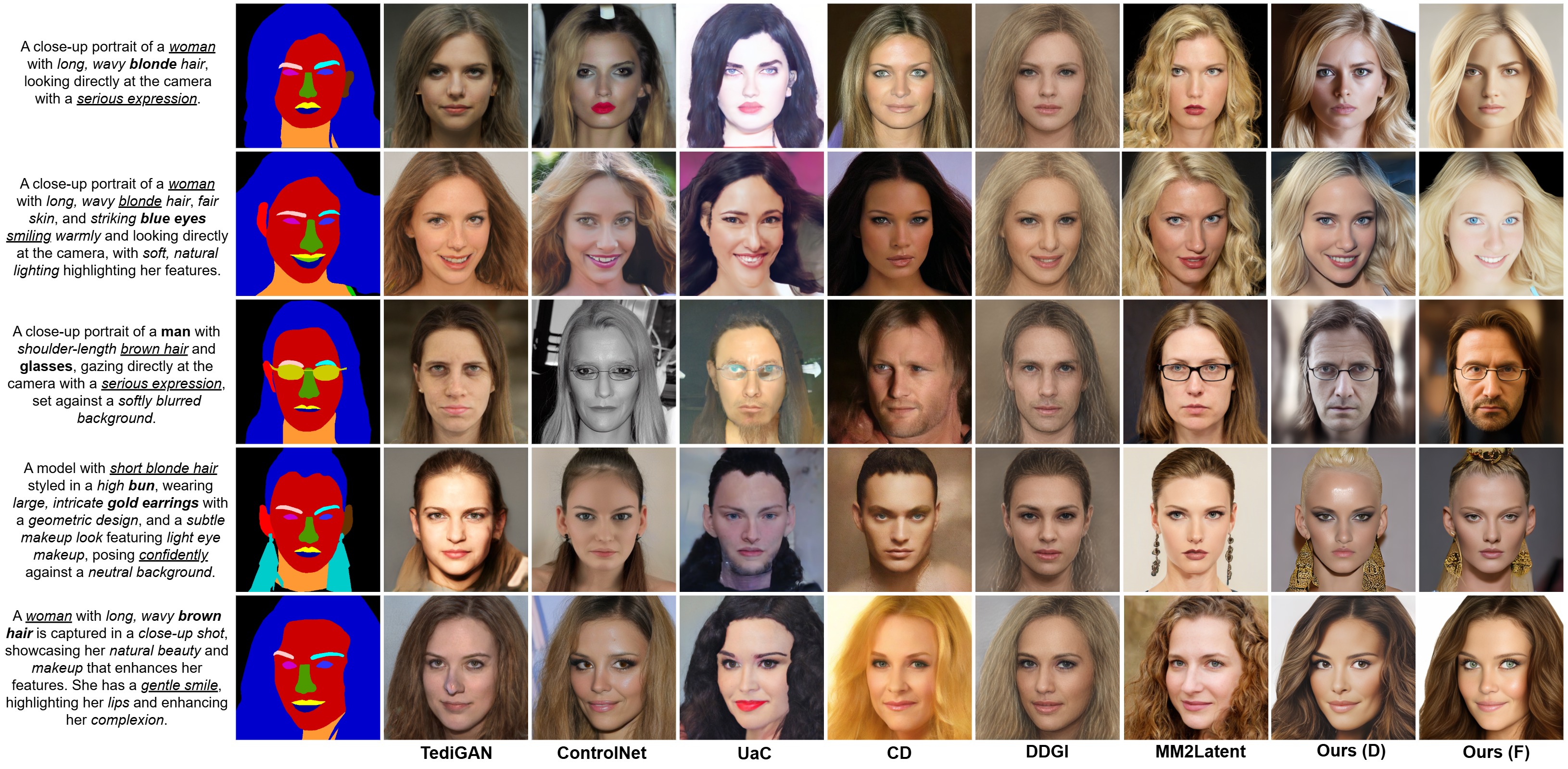}
\caption{\textbf{Qualitative comparison of text-and-mask-conditioned face generation.} Our MMFace-DiT excels at integrating complex textual attributes (e.g., the fourth row with \textit{gold earrings} and \textit{high bun}) with the mask's structure, producing photorealistic and coherent outputs where prior methods introduce artifacts or fail to adhere to the prompt.}
\label{fig:Masks}
\end{figure*}

\subsection{The Dual-Stream MMFace-DiT Block}
The core of our architecture is the Dual-Stream MMFace-DiT block, detailed in Figure~\ref{fig:block}. It processes image ($T_i$) and text ($T_t$) tokens through parallel streams that are deeply and continuously fused. The block's workflow is governed by three key mechanisms: an \textit{Adaptive Layer Normalization (AdaLN)} scheme for fine-grained conditioning, a shared \textit{RoPE Attention} layer for central fusion, and \textit{Gated Residual Connections} (Gate) for dynamically balancing information flow from the attention and subsequent MLP layers. Each MLP is a two-layer feed-forward network that expands the hidden dimension by a factor of 4, with a GeLU activation~\cite{hendrycks2016gaussian} between the two linear layers.

\vspace{-1em}

\paragraph{Adaptive Layer Normalization (AdaLN).} The global conditioning vector $C_{\text{global}}$ orchestrates the behavior of each block. It is transformed by a linear layer to generate a comprehensive set of modulation parameters $\{\gamma, \beta, \alpha\}$ for the attention and MLP components of both streams independently. This allows the text, timestep, and active modality to exert fine-grained, layer-specific control over the entire network.

\vspace{-1em}
\paragraph{Shared RoPE Attention for Deep Fusion.} The central fusion mechanism is a single, shared multi-head attention operation. Tokens from both streams are projected into query, key, and value tensors and concatenated into unified representations: $Q = [Q_i; Q_t]$, $K = [K_i; K_t]$, and $V = [V_i; V_t]$. We apply \textbf{Rotary Position Embeddings (RoPE)} to the combined query and key tensors. Specifically, image tokens receive \textbf{2D axial RoPE} encoding (capturing spatial relationships across height and width), while text tokens receive \textbf{1D sequential RoPE} encoding. This hybrid approach naturally handles the heterogeneous structure of 2D image patches and 1D text tokens within a single unified attention operation:
\begin{equation}
\text{Attention}(Q, K, V) = \text{softmax}\left(\frac{\text{RoPE}(Q) \text{RoPE}(K)^T}{\sqrt{d_k}}\right)V
\end{equation}
This allows every image patch to attend to every text token and vice-versa, enabling a deep, bidirectional flow of information essential for \textit{precise semantic alignment and spatial fidelity}.

\vspace{-1em}

\paragraph{Gate.} Following attention and MLP operations, we employ Gate to modulate their outputs. For an input stream $T_{\text{in}}$ and a block operation $F(\cdot)$, the update is:
\begin{equation}
T_{\text{out}} = T_{\text{in}} + \alpha \odot F(\text{AdaLN}(T_{\text{in}}, \gamma, \beta))
\end{equation}
The gating scalar $\alpha$, derived from $C_{\text{global}}$, acts as a dynamic, learned filter. It allows the network to selectively emphasize or suppress information flow from specific modalities, crucial for preventing one strong modality (e.g., a dense sketch) from overpowering subtle semantic cues from the text.

\begin{table*}[t]
\centering
\small
\renewcommand{\arraystretch}{1}
\sisetup{detect-weight, mode=text, table-number-alignment=center}
\setlength{\tabcolsep}{10pt}
\sisetup{detect-weight, mode=text}
\begin{tabular}{@{} l
                S[table-format=2.2]
                S[table-format=1.2]
                S[table-format=1.2]
                S[table-format=2.2]
                S[table-format=2.2]
                S[table-format=2.2]
                S[table-format=1.2]
                S[table-format=1.4] @{}}
\toprule
\textbf{Method} & {\textbf{FID} $\downarrow$} & {\textbf{LPIPS} $\downarrow$} & {\textbf{SSIM} $\uparrow$} & {\textbf{ACC} $\uparrow$} & {\textbf{mIoU} $\uparrow$} & {\textbf{CLIP} $\uparrow$} & {\textbf{Dist.} $\downarrow$} & {\textbf{LLM Sc.} $\uparrow$} \\
\midrule
TediGAN~\cite{xia2021tedigan} & 62.55 & 0.43 & 0.48 & 79.77 & 39.02 & 25.26 & 0.75 & 0.4061 \\
ControlNet~\cite{zhang2023adding} & 49.39 & 0.57 & 0.41 & 82.86 & 43.95 & 25.39 & 0.75 & 0.3103 \\
UAC~\cite{nair2023unite} & 48.88 & 0.46 & 0.48 & 78.27 & 38.82 & 23.75 & 0.76 & 0.3516 \\
CD~\cite{huang2023collaborative} & 49.00 & 0.56 & 0.46 & 85.69 & 38.85 & 25.07 & 0.75 & 0.3029 \\
DDGI~\cite{kim2024diffusion} & 50.88 & 0.45 & 0.49 & 86.00 & 36.02 & 24.29 & 0.76 & 0.3851 \\
MM2Latent~\cite{meng2024mm2latent} & 49.78 & 0.59 & 0.45 & 84.57 & 38.19 & 26.78 & 0.73 & 0.3619 \\
\midrule
\rowcolor[HTML]{E8F5E9}
\textbf{Ours (D)} & 27.95 & 0.34 & 0.51 & \textbf{93.95} & 49.16 & \textbf{31.69} & \textbf{0.68} & 0.6006 \\
\rowcolor[HTML]{FFF3F3}
\textbf{Ours (F)} & \textbf{16.63} & \textbf{0.34} & \textbf{0.53} & 93.74 & \textbf{50.12} & 31.34 & 0.69 & \textbf{0.6372} \\
\bottomrule
\end{tabular}
\caption{\textbf{Quantitative results for Text + Mask conditioned face generation.} 
Our MMFace-DiT variants include \textbf{Ours (D)}, trained with diffusion-based DDPM objectives, and \textbf{Ours (F)}, trained using flow-matching objectives. 
Both substantially outperform all baselines across perceptual quality and text-image alignment metrics. Best results are in \textbf{bold}.}
\label{tab:mask_results}
\end{table*}

\subsection{Training Objectives and Optimization}
Our model supports two complementary, diffusion-based training paradigms, which we explore to optimize performance.

\vspace{-1em}

\paragraph{1. DDPM with Min-SNR Weighting.} We train our model $\epsilon_\theta$ to predict the noise $\epsilon$ added to a latent $z_0$ at timestep $t$. To accelerate convergence and improve perceptual quality at high resolutions, we adopt the Min-SNR weighting strategy~\cite{hang2023efficient}, which balances the MSE loss contribution across varying noise levels:
{\small
\begin{equation}
\mathcal{L}_{\text{DDPM}} = \mathbb{E}_{t, z_0, c_{\text{sp}}, m, p, \epsilon} \left[ w(t) \left\| \epsilon - \epsilon_\theta(z_t, t, z_c, m, C_{\text{global}}, c_{\text{seq}}) \right\|^2 \right]
\label{eq:loss_ddpm}
\end{equation}
}
where $w(t) = \min(\text{SNR}(t), \lambda) / \text{SNR}(t)$, with $\lambda = 5.0$.

\vspace{-1em}

\paragraph{2. Rectified Flow Matching (RFM).} As an alternative, we also adopt the widely popular Rectified Flow Matching paradigm~\cite{liu2022flow, esser2024scaling, wu2025qwen}, which treats diffusion as learning a velocity field between noise ($x_0$) and data ($x_1$). We sample a continuous time $t \sim \mathcal{U}[0, 1]$ and construct an interpolated latent $z_t = (1-t)x_0 + tx_1$. The model predicts the constant velocity $v = x_1 - x_0$:
{\small
\begin{equation}
\mathcal{L}_{\text{RFM}} = \mathbb{E}_{t, x_0, x_1, c_{\text{sp}}, m, p} \left[ \left\| v_\theta(z_t, t, z_c, m, C_{\text{global}}, c_{\text{seq}}) - (x_1 - x_0) \right\|^2 \right]
\label{eq:loss_rfm}
\end{equation}
}
This formulation eliminates the need for variance schedules and has shown improved stability in high-resolution synthesis with faster inference times~\cite{lipman2022flow, cai2025hidream}.

\vspace{-1em}

\paragraph{Implementation and Resource-Efficient Training.}
Our MMFace-DiT model has 1.345B parameters, 28 transformer blocks (as in DiT-XL~\cite{DiT2023}), a hidden size of 1152, and 16 attention heads. It operates on a 32-channel latent input ($z_t$ + $z_c$) from the 16-channel FLUX VAE, featuring shared 2D RoPE fusion and a dynamic Modality Embedder for flexible conditioning. The model is trained progressively using either DDPM or RFM objectives, first at $256^2$ (300 epochs; batch 32; lr=$10^{-4}$) and then fine-tuned at $512^2$ (50 epochs; effective batch 16 via gradient accumulation; lr=$10^{-6}$). Despite its scale, training was highly efficient through aggressive memory optimizations, including \texttt{bfloat16} precision, 8-bit AdamW, full gradient checkpointing, and precomputed VAE latents. The entire model was trained on a modest setup of just \textit{two NVIDIA RTX 5000 Ada GPUs}, demonstrating that MMFace-DiT can operate in resource-constrained environments. Comprehensive details of all hyperparameters, architectural specifics, and training schedules is provided in the Supplementary Material.

%% file: sec/4_experiments.tex
\section{Experiments}
\label{sec:experiments}


\begin{figure*}[ht]
    \centering
    \includegraphics[width=0.95\textwidth]{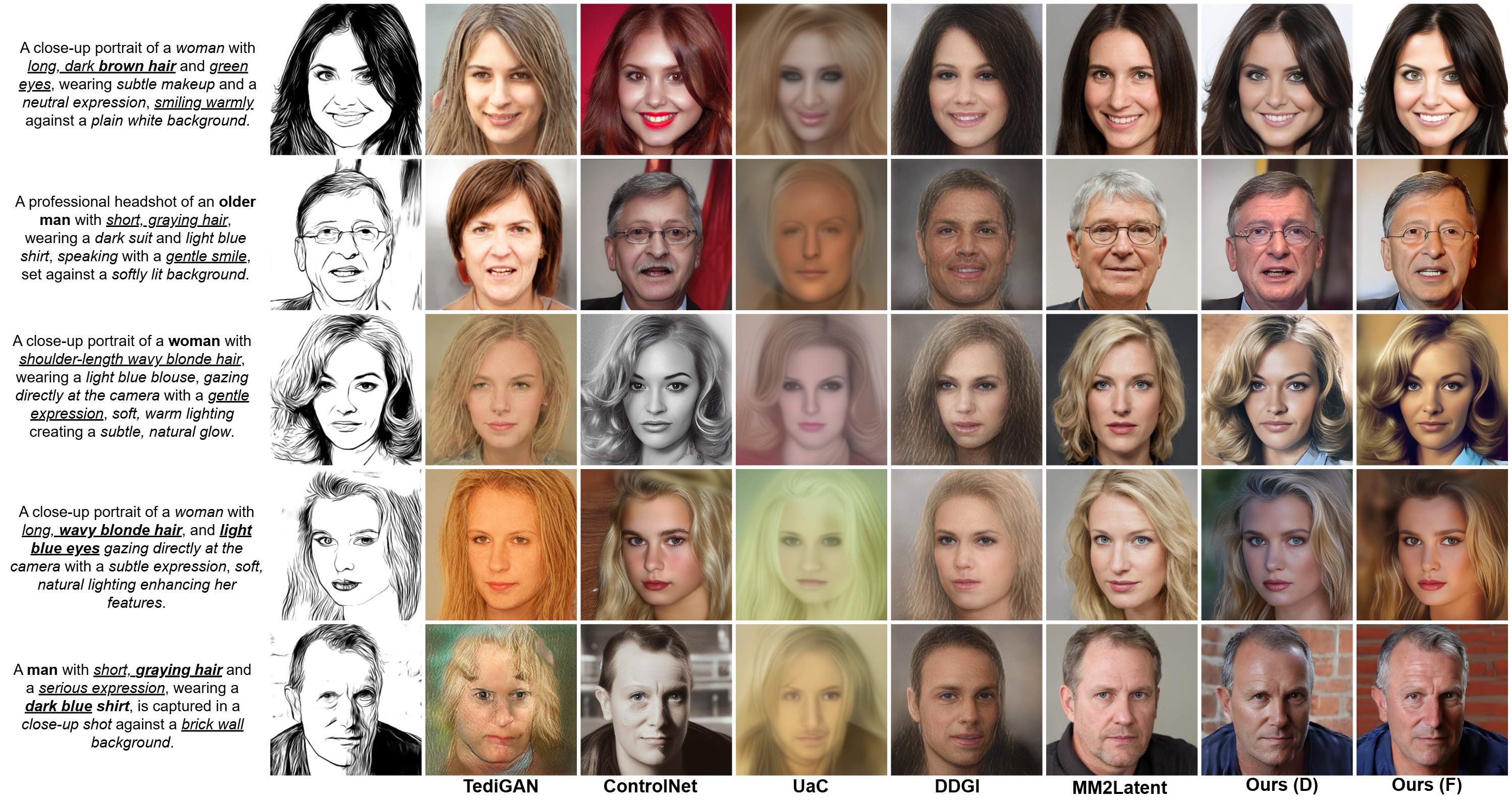}
    \caption{\textbf{Qualitative comparison of text-and-sketch-conditioned face generation.} Our MMFace-DiT excels at balancing the strong structural priors of a sketch with nuanced textual descriptions (e.g., the fifth row with \textit{graying hair} and \textit{dark blue shirt}), producing photorealistic outputs with high textural fidelity while prior methods introduce color artifacts, blur, or semantic inconsistencies.}
    \label{fig:Baseline}
\end{figure*}

\subsection{Evaluation protocol}

\paragraph{Datasets.}
We conduct our training on a combined dataset of \textbf{CelebA-HQ}~\cite{xia2021tedigan} and \textbf{FFHQ}~\cite{karras2019style}. Since FFHQ lacks official annotations, we generate both spatial conditions: \textit{semantic masks} using a pre-trained Segformer face-parsing model~\cite{xie2021segformer}, and \textit{sketches} via the U2Net model~\cite{qin2020u2}. To ensure rich textual conditioning, we further construct a VLM-based captioning pipeline~\cite{zhu2025internvl3} that produces multiple diverse and semantically detailed captions per image across both datasets.

\vspace{-1em}


\paragraph{Baselines.} 
For mask-conditioning, we compare against six leading approaches: \textbf{TediGAN}~\cite{xia2021tedigan}, \textbf{ControlNet}~\cite{zhang2023adding}, \textbf{Unite and Conquer (UaC)}~\cite{nair2023unite}, \textbf{Collaborative Diffusion (CD)}~\cite{huang2023collaborative}, and \textbf{DDGI}~\cite{kim2024diffusion}. We use official public weights where available and faithfully re-implemented DDGI as its code was not public. For sketch-conditioning, we adopt the same baselines but exclude CD, which lacks pre-trained weights for the task.


\vspace{-1em}

\paragraph{Evaluation Metrics.}
We assess performance using a comprehensive suite of metrics. \textbf{Image realism} is measured by Fréchet Inception Distance (FID)~\cite{heusel2017gans} and Learned Perceptual Image Patch Similarity (LPIPS)~\cite{zhang2018unreasonable}. For masks, we evaluate \textbf{structural integrity} with Pixel Accuracy (ACC) and mean Intersection-over-Union (mIoU). \textbf{Spatial fidelity} is further assessed with the multi-scale Structural Similarity Index Measure (SSIM)~\cite{wang2004image}. Finally, we quantify \textbf{text-image alignment} using CLIP Score and Distance~\cite{radford2021learning, hessel2021clipscore}, and capture more nuanced semantic consistency with an LLM Score (LLM Sc.)~\cite{lu2023llmscore}.

\subsection{Qualitative Analysis}
\paragraph{Mask-Conditioned Generation.}
As shown in \cref{fig:Masks}, our MMFace-DiT (Ours (D), trained with diffusion-based DDPM objectives, and Ours (F), trained using flow-matching objectives.) achieves superior photorealism and precise multimodal alignment. Competing methods often introduce artifacts, miss attributes, or degrade identity coherence, whereas ours faithfully renders complex descriptors like \textit{wavy blonde hair} and \textit{blue eyes} while maintaining structural fidelity. Notably, our method excels at reproducing intricate attributes such as a \textit{high bun} or \textit{gold earrings} with accurate geometry and material realism. This \textit{superiority} stems from our dual-stream design, where shared RoPE attention prevents modal dominance by facilitating a deep, token-level interaction, ensuring fine-grained integration of text and mask cues.


\vspace{-1em}

\paragraph{Sketch-Conditioned Generation.}
As illustrated in \cref{fig:Baseline}, our approach also excels in sketch-conditioned synthesis, producing lifelike faces that adhere to both modalities. While all baselines frequently generate oversmoothed or semantically inconsistent outputs, our method preserves detailed geometry and natural skin texture. Fine-grained attributes from the text—from expressions like \textit{smiling warmly} to details like a \textit{dark blue shirt}—are rendered with realistic shading and tone. This is a direct result of our gated residual connections, which dynamically balance the strong geometric priors from the sketch against the subtle, semantic cues from the text, ensuring structural fidelity without sacrificing photorealism.

\begin{table}[t]
\centering
\small
\scriptsize
\renewcommand{\arraystretch}{1}
\sisetup{detect-weight, mode=text, table-number-alignment=center}
\setlength{\tabcolsep}{4.5pt}
\begin{tabular}{@{} l
                S[table-format=3.2]
                S[table-format=1.2]
                S[table-format=1.2]
                S[table-format=2.2]
                S[table-format=1.2]
                 S[table-format=1.2]
                 @{}}
\toprule
\textbf{Method} & {\textbf{FID} $\downarrow$} & {\textbf{LPIPS} $\downarrow$} & {\textbf{SSIM} $\uparrow$} & {\textbf{CLIP} $\uparrow$} & {\textbf{Dist.} $\downarrow$} & {\textbf{LLM Sc.} $\uparrow$} \\
\midrule
TediGAN~\cite{xia2021tedigan}      & 121.24 & 0.55 & 0.30 & 21.62 & 0.78 & 0.10 \\
ControlNet~\cite{zhang2023adding}   & 67.13  & 0.54 & 0.56 & 26.17 & 0.74 & 0.44 \\
UAC~\cite{nair2023unite}          & 118.52 & 0.61 & 0.41 & 22.92 & 0.77 & 0.27 \\
DDGI~\cite{kim2024diffusion}         & 56.57  & 0.43 & 0.51 & 23.95 & 0.76 & 0.43 \\
MM2Latent~\cite{meng2024mm2latent} & 40.91 & 0.58 & 0.46 & 27.04 & 0.73 & 0.39 \\
\midrule
\rowcolor[HTML]{E8F5E9}
\textbf{Ours (D)} & 27.67 & 0.24 & \textbf{0.72} & \textbf{31.56} & \textbf{0.68} & 0.69 \\
\rowcolor[HTML]{FFF3F3}
\textbf{Ours (F)} & \textbf{9.14} & \textbf{0.20} & 0.70 & 31.30 & 0.69 & \textbf{0.72} \\
\bottomrule
\end{tabular}
\caption{\textbf{Quantitative results for Text + Sketch conditioned face generation.} 
Our MMFace-DiT includes \textbf{Ours (D)} (diffusion-based DDPM) and \textbf{Ours (F)} (flow-matching) variants.}
\label{tab:sketch_results}
\end{table}

\subsection{Quantitative Results}
\paragraph{Text and Mask Conditioning.}
Table~\ref{tab:mask_results} shows our model substantially improves perceptual quality and semantic alignment. Our diffusion-trained model (D) attains an FID of \textbf{27.95}, a \textbf{42.8\%} reduction relative to the strongest baseline, UAC. This is complemented by major gains over other leading methods, including a \textbf{24.8\%} higher CLIP Score than ControlNet and a \textbf{24.4\%} drop in LPIPS versus DDGI. While both our training variants excel, our flow-matching model (F) further pushes the state-of-the-art, reducing the FID by a relative 40.5\% to \textbf{16.63}. We \textit{attribute} these gains to our core design: (i) the dual-stream architecture prevents modal dominance, (ii) shared RoPE attention enables dense, bidirectional fusion, and (iii) our Modality Embedder and gating mechanisms provide adaptive control.

\vspace{-1em}

\paragraph{Text and Sketch Conditioning.}
Improvements are even more pronounced for sketch conditioning, as shown in Table~\ref{tab:sketch_results}. Our diffusion model (D) achieves an FID of \textbf{27.67}, a remarkable \textbf{32.4\%} improvement over the strongest prior method, MM2Latent, along with major gains across all key metrics, including a \textbf{44.2\%} lower LPIPS than DDGI and a \textbf{56.8\%} higher LLM semantic-consistency score over ControlNet. Our flow-matching variant (F) establishes a new benchmark over (D), reducing FID by 66.9\%. These results reflect our \textit{model's ability} to (i) fuse local sketch geometry with global textual cues via shared attention and gating, (ii) dynamically adapt to the sketch modality through the Modality Embedder, and (iii) leverage our VLM-augmented captions for richer text-visual correspondence.

\begin{figure*}[ht]
\centering
\includegraphics[width=0.85\textwidth]{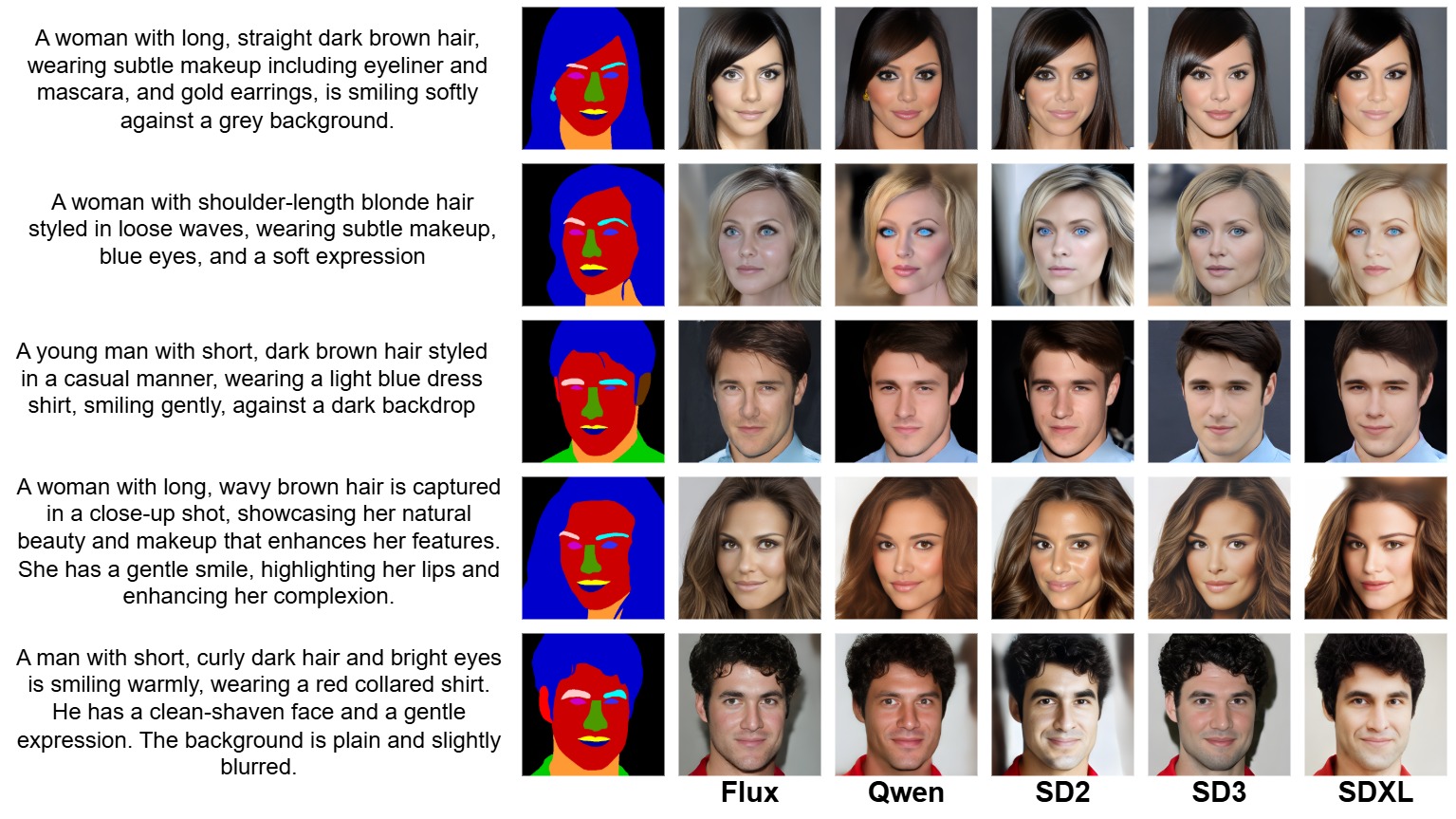}
\caption{\textbf{Qualitative comparison of VAE backbones integrated into MMFace-DiT.} The choice of VAE highlights a clear trade-off between statistical fidelity and perceptual quality. While some VAEs introduce visual artifacts such as oversaturation (SD3) or color desaturation (SDXL), Flux consistently yields the most perceptually faithful outputs with superior color accuracy and natural texture.}
\label{fig:VAE_Ablations_2}
\end{figure*}

%% file: sec/5_Ablation.tex
\section{Ablation Study}
\paragraph{Ablation on model components.}
We conducted a systematic ablation to quantify the contribution of each component in \textbf{MMFace-DiT} (Table~\ref{tab:component_ablation}). The baseline \textbf{DiT} (Model-1) achieves an FID of 44.52 and mIoU of 44.86 but requires separate training per spatial modality. Introducing the \textbf{Modality Embedder (ME)} (Model-2) enables shared spatial learning, improving FID by \textbf{9.1\%} and mIoU by \textbf{3.3\%}. Adding the \textbf{Dual-Stream (DS)} design (Model-3) jointly enhances semantic and spatial alignment, increasing the CLIP Score by \textbf{22.1\%} (24.53$\rightarrow$29.69) and mIoU by \textbf{5.5\%} (46.34$\rightarrow$48.91). Replacing standard attention with \textbf{RoPE attention} (Model-4) further strengthens multimodal fusion, yielding the highest mIoU (50.05) and improved CLIP (31.42). Finally, replacing SD2 with the \textbf{Flux VAE} (Model-5) provides the best overall balance, reducing FID by an additional \textbf{17.2\%} while achieving the highest SSIM (0.51) and Pixel Accuracy (93.95\%), resulting in strong perceptual realism with preserved spatial fidelity.


\vspace{-1em}


\begin{table}[h!]
\centering
\scriptsize
\renewcommand{\arraystretch}{1}
\sisetup{detect-weight, mode=text, table-number-alignment=center}
\setlength{\tabcolsep}{2.2pt}
\begin{tabular}{@{}l ccc c ccc ccc@{}}
\toprule
& \multicolumn{4}{c}{\textbf{Components}} & \multicolumn{3}{c}{\textbf{Semantic Metrics}} & \multicolumn{3}{c}{\textbf{Spatial Metrics}} \\
\cmidrule(lr){2-5} \cmidrule(lr){6-8} \cmidrule(lr){9-11}
\textbf{Model} & ME & DS & RoPE & VAE & FID $\downarrow$ & LPIPS $\downarrow$ & CLIP $\uparrow$ & SSIM $\uparrow$ & ACC $\uparrow$ & mIoU $\uparrow$ \\
\midrule
Model-1 & $\times$   & $\times$   & $\times$   & SD2   & 44.52          & 0.486          & 24.53          & 0.44  & 86.65 & 44.86 \\
Model-2 & \checkmark & $\times$   & $\times$   & SD2   & 40.49          & 0.366          & 24.31          & 0.46  & 87.91 & 46.34 \\
Model-3 & \checkmark & \checkmark & $\times$   & SD2   & 35.61          & 0.367          & 29.69  
       & 0.49  & 90.79 & 48.91 \\
Model-4 & \checkmark & \checkmark & \checkmark & SD2   & 33.77          & \textbf{0.326}          & 31.42          & 0.50  & 92.29 & \textbf{50.05} \\
Model-5 & \checkmark & \checkmark & \checkmark & Flux  & \textbf{27.95} & 0.340 & \textbf{31.69} & \textbf{0.51} & \textbf{93.95} & 49.16 \\
\bottomrule
\end{tabular}
\caption{\textbf{Ablation study on core model components with spatial metrics.} We incrementally add our innovations, demonstrating the impact of the Modality Embedder (ME), Dual-Stream (DS) design, Rotary Position Embedding (RoPE) Attention, and the final VAE choice. Spatial metrics (SSIM, ACC, mIoU) show simultaneous improvement with semantic metrics, proving mitigation of modality dominance.}
\label{tab:component_ablation}
\end{table}


\paragraph{Ablation on VAE Architectures.}
To select the optimal latent space representation, we evaluated five frozen VAEs: SD2~\cite{rombach2022high}, SDXL~\cite{podell2023sdxl}, Qwen-Image~\cite{wu2025qwen}, SD3~\cite{esser2024scaling}, and Flux-Krea~\cite{flux1kreadev2025}, with results summarized in Table~\ref{tab:vae_ablation}. The SD2 baseline exhibits limited chromatic diversity. SDXL shows a modest 3.4\% relative LPIPS improvement over SD2, while Qwen optimized for text-in-image alignment underperforms both. Notably, all three models often fail to reproduce vivid attributes such as “blue eyes” (Fig.~\ref{fig:VAE_Ablations_2}, row 2). Although SD3 delivers the lowest FID, its outputs appear glossy and less photorealistic due to exaggerated specular highlights. In contrast, \textbf{Flux} attains the best perceptual quality (LPIPS 0.239 on sketches), owing to its training on curated, photorealistic data~\cite{lee2025flux} that yields a latent space preserving both skin texture and color fidelity, making it the most balanced and artifact-free VAE backbone.


\begin{table}[t]
\centering
\small
\scriptsize
\renewcommand{\arraystretch}{1}
\sisetup{detect-weight, mode=text, table-number-alignment=center}
\setlength{\tabcolsep}{5pt}
\begin{tabular}{@{} l
                S[table-format=2.0]
                S[table-format=1.4]
                S[table-format=2.2]
                S[table-format=1.3]
                S[table-format=2.2]
                S[table-format=1.3] @{}}
\toprule
\textbf{VAE} & {\textbf{Dim.}} & {\textbf{Scaling}} & {\textbf{FID (M)}$\downarrow$} & {\textbf{LPIPS (M)}$\downarrow$} & {\textbf{FID (S)}$\downarrow$} & {\textbf{LPIPS (S)}$\downarrow$} \\
\midrule
SD2   & 4  & 0.1821 & 33.77 & 0.349 & 32.58 & 0.245 \\
SDXL  & 4  & 0.1821 & 23.00 & \textbf{0.337} & 24.16 & 0.258 \\
Qwen  & 16 & {--}   & 33.09 & 0.359 & 30.16 & 0.251 \\
SD3   & 16 & 1.5305 & \textbf{20.73} & 0.368 & \textbf{17.13} & 0.251 \\
Flux & 16 & 0.3611 & 27.95 & 0.340 & 27.67 & \textbf{0.239} \\
\bottomrule
\end{tabular}
\caption{\textbf{Ablation study on VAE architectures.} We evaluate the reconstruction and generative quality of our model using five different pre-trained VAEs. (M) refers to mask-conditioned and (S) to sketch-conditioned generation.}
\label{tab:vae_ablation}
\end{table}

%% file: sec/6_conclusion.tex
\section{Conclusion}
\label{sec:conclusion}
In this paper, we introduced MMFace-DiT, a novel, unified architecture for high-fidelity face generation from text, masks, and sketches. Its dual-stream blocks with shared RoPE attention enable deep cross-modal fusion, while the dynamic modality embedder allows flexible conditioning within a single model. By providing a scalable, end-to-end alternative to brittle compositional pipelines, MMFace-DiT attains state-of-the-art performance, outperforming five leading methods and defining a new benchmark for unified multi-modal face synthesis.


%% file: sec/X_suppl.tex
\clearpage
\setcounter{page}{1}
\setcounter{table}{0}
\setcounter{figure}{0}
\setcounter{section}{0}
\maketitlesupplementary

\section{Implementation Details}
Our Dual-Stream MMFace-DiT follows a DiT-XL~\citesupp{DiT2023x} configuration with 1.345 billion parameters, including 28 transformer blocks, a hidden size of 1152, and 16 attention heads. The model operates with a patch size of 2 in the latent space of the 16-channel FLUX VAE, forming a 32-channel input tensor (16 for the noisy latent and 16 for the spatial condition). Key to our architecture is a shared attention mechanism implementing \textit{2D Rotary Position Embedding (RoPE)}, which uses a base period ($\theta$ = 10,000) to encode positional information in the query and key tensors of both streams. Dynamic conditioning is driven by a \textit{Modality Embedder}, a learnable lookup layer that converts the discrete spatial condition (mask or sketch) into a dense vector. This vector enriches the global conditioning signal, from which our AdaLN scheme derives twelve distinct parameter vectors per block to control the shift, scale, and gate of both the attention and MLP sub-layers. Training proceeds progressively using either a DDPM objective with Min-SNR weighting or a Rectified Flow Matching (RFM) objective. We first train from scratch at $256 \times 256$ resolution for 300 epochs with a batch size of 32 per GPU, a learning rate of $1 \times 10^{-4}$, then fine-tune at $512 \times 512$ for 50 epochs with an effective batch size of 16 (via 2-step gradient accumulation) per GPU and a learning rate of $1 \times 10^{-6}$. To enable classifier-free guidance, we apply a 5\% dropout probability to both the text and spatial conditioning inputs during training. We employ the 8-bit AdamW optimizer with a cosine learning rate scheduler and ensure memory efficiency using \texttt{bfloat16} mixed precision, full gradient checkpointing, and an Exponential Moving Average (EMA) of model weights with a decay of 0.9999. The complete details for all the hyperparameters are provided in Table~\ref{tab:hyperparameters}.

\begin{algorithm}[!h]
\caption{MMFace-DiT Training with Diffusion Objective}
\label{alg:ddpm_training}
\begin{algorithmic}[1]
\State \textbf{Input:} DiT model $\epsilon_\theta$, VAE Encoder $\mathcal{E}_{vae}$, CLIP Text Encoder $\mathcal{E}_{text}$, Noise Scheduler $\mathcal{S}$
\State \textbf{Require:} Dataset $D$ of triplets $(x, c_{sp}, p)$ and modality flag $m$
\Repeat
    \State Sample a batch $(x, c_{sp}, p, m)$ from $D$
    \State $z_0 \gets \mathcal{E}_{vae}(x)$; \quad $z_c \gets \mathcal{E}_{vae}(c_{sp})$ \Comment{Encode images to latents}
    \State $c_{pooled}, c_{seq} \gets \mathcal{E}_{text}(p)$ \Comment{Encode text prompt}
    \State Sample timestep $t \sim \mathcal{U}\{1, ..., T\}$ and noise $\epsilon \sim \mathcal{N}(0, \mathbf{I})$
    \State $z_t \gets \mathcal{S}.\text{add\_noise}(z_0, \epsilon, t)$ \Comment{Apply forward diffusion process}
    \State $z_{in} \gets \text{concat}(z_t, z_c)$ \Comment{Concatenate noisy latent and spatial condition}
    \State Form global conditioning $C_{global}$ from $t, c_{pooled}, m$
    \State $\epsilon_{pred} \gets \epsilon_\theta(z_{in}, t, C_{global}, c_{seq})$ \Comment{Predict noise from all conditions}
    \State Calculate Min-SNR weight $w(t)$ based on scheduler alphas at timestep $t$
    \State $\mathcal{L} \gets w(t) \cdot \mathbb{E}[\|\epsilon - \epsilon_{pred}\|^2]$ \Comment{Compute weighted MSE loss}
    \State Update model parameters $\theta$ using gradient descent on $\mathcal{L}$
\Until{converged}
\end{algorithmic}
\end{algorithm}

\begin{table*}[ht]
\centering
\small
\renewcommand{\arraystretch}{1.2}
\setlength{\tabcolsep}{10pt}
\begin{tabular}{l l l}
\toprule
\textbf{Hyperparameter} & \textbf{Value (Stage 1: 256x256)} & \textbf{Value (Stage 2: 512x512 Fine-tuning)} \\
\midrule
\multicolumn{3}{l}{\textit{\textbf{Model Architecture}}} \\
\quad Hidden Size (D) & 1152 & 1152 \\
\quad Depth (Transformer Blocks) & 28 & 28 \\
\quad Attention Heads & 16 & 16 \\
\quad Patch Size & 2 & 2 \\
\quad Input Channels & 32 (16 for latent, 16 for condition) & 32 (16 for latent, 16 for condition) \\
\quad MLP Ratio & 4.0 & 4.0 \\
\quad RoPE Theta ($\theta$) & 10,000 & 10,000 \\
\midrule
\multicolumn{3}{l}{\textit{\textbf{External Components}}} \\
\quad VAE & FLUX VAE (16-channel) & FLUX VAE (16-channel) \\
\quad Text Encoder & CLIP (from SD 2.1-base) & CLIP (from SD 2.1-base) \\
\midrule
\multicolumn{3}{l}{\textit{\textbf{Training Strategy}}} \\
\quad Training Objectives & DDPM (Min-SNR) / RFM & DDPM (Min-SNR) / RFM \\
\quad Epochs & 300 & 50 \\
\quad Steps & 440700 & 146900 \\
\quad Batch Size (per GPU) & 32 & 8 \\
\quad Gradient Accumulation Steps & 1 & 2 \\
\quad \textbf{Effective Batch Size} & \boldmath$32 \times 2$ & \boldmath$8 \times 2 \times 2$ \\
\midrule
\multicolumn{3}{l}{\textit{\textbf{Optimizer (8-bit AdamW)}}} \\
\quad Learning Rate & $1 \times 10^{-4}$ & $1 \times 10^{-6}$ \\
\quad LR Scheduler & Cosine Decay & Cosine Decay \\
\quad LR Warmup Steps & 200 & 100 \\
\quad Adam $\beta_1$ & 0.9 & 0.9 \\
\quad Adam $\beta_2$ & 0.999 & 0.999 \\
\quad Weight Decay & 0.01 & 0.01 \\
\quad Adam $\epsilon$ & $1 \times 10^{-8}$ & $1 \times 10^{-8}$ \\
\midrule
\multicolumn{3}{l}{\textit{\textbf{Regularization \& Efficiency}}} \\
\quad Max Gradient Norm & 0.5 & 0.5 \\
\quad Conditioning Dropout & 5\% (Text, Mask, Sketch) & 5\% (Text, Mask, Sketch) \\
\quad EMA Decay & 0.9999 & 0.9999 \\
\quad Mixed Precision & `bfloat16` & `bfloat16` \\
\quad Gradient Checkpointing & Enabled & Enabled \\
\quad Dataloader Workers & 4 & 4 \\
\bottomrule
\end{tabular}
\caption{Detailed hyperparameters for the two-stage progressive training of our MMFace-DiT model. Stage 1 trains the model from scratch at 256x256 resolution, and Stage 2 fine-tunes the resulting checkpoint at 512x512 resolution.}
\label{tab:hyperparameters}
\end{table*}

\section{Training Objectives and Inference}
Our MMFace-DiT architecture is compatible with two distinct generative training paradigms: Denoising Diffusion Probabilistic Modeling (DDPM)~\citesupp{ho2020denoisingx} and Rectified Flow Matching (RFM)~\citesupp{liu2022flowx}. Both approaches leverage the same core model and conditioning mechanisms but differ in their optimization objective and sampling procedure. To rigorously evaluate the efficacy and generalization of our method, all inference and sampling procedures described herein are conducted exclusively on the official CelebA-HQ test split of $6,000$ images, a held-out partition on which our model was not trained.

\subsection{DDPM Objective}
Under the DDPM framework, the model $\epsilon_\theta$ is trained to predict the noise $\epsilon$ added to a clean latent $z_0$ at a given timestep $t$. The model is conditioned on a concatenated input tensor $z_{in}$ (containing the noised latent $z_t$ and the spatial condition latent $z_c$), a global conditioning signal $C_{global}$ (derived from the timestep $t$, pooled text embedding $c_{pooled}$, and modality flag $m$), and the text sequence embedding $c_{seq}$. We use a Min-SNR weighting strategy~\citesupp{hang2023efficientx} to stabilize training by re-weighting the loss at each timestep. The complete training and inference procedures, which employ the efficient DPM-Solver Multistep scheduler~\citesupp{lu2022dpmx} with Classifier-Free Guidance (CFG)~\citesupp{ho2022classifierx}, are detailed in \cref{alg:ddpm_training} and \cref{alg:ddpm_inference}, respectively.

\subsection{RFM Objective}
In the RFM paradigm, the model $v_\theta$ learns to predict the constant-velocity vector $v = z_1 - z_0$ that connects a noise sample $z_0 \sim \mathcal{N}(0, \mathbf{I})$ to a data sample $z_1$. The model is trained on an interpolated latent $z_t = (1-t)z_0 + t z_1$ and receives the same set of conditioning inputs as in the DDPM setup ($z_c$, $C_{global}$, and $c_{seq}$) to predict the target velocity. Inference involves integrating the predicted velocity field from $t=0$ to $t=1$ using an ODE solver like the Euler method, also guided by CFG. The specific training and inference steps are outlined in \cref{alg:rfm_training} and \cref{alg:rfm_inference}.

\begin{algorithm}[ht]
\caption{MMFace-DiT Inference with Diffusion Objective}
\label{alg:ddpm_inference}
\begin{algorithmic}[1]
\State \textbf{Input:} Prompt $p$, spatial condition $c_{sp}$, modality flag $m$, guidance scale $\omega$
\State \textbf{Require:} Trained DiT model $\epsilon_\theta$, VAE Encoder/Decoder $\mathcal{E}_{vae}, \mathcal{D}_{vae}$, CLIP Encoder $\mathcal{E}_{text}$, Scheduler $\mathcal{S}$
\State $z_c \gets \mathcal{E}_{vae}(c_{sp})$ \Comment{Encode spatial condition}
\State $c_{pooled}^{cond}, c_{seq}^{cond} \gets \mathcal{E}_{text}(p)$ \Comment{Encode conditional prompt}
\State $c_{pooled}^{uncond}, c_{seq}^{uncond} \gets \mathcal{E}_{text}(\text{" "})$ \Comment{Encode unconditional (null) prompt}
\State Sample initial latent $z_T \sim \mathcal{N}(0, \mathbf{I})$
\State Set scheduler timesteps $\mathcal{T} = \{T, T-1, ..., 1\}$
\For{$t$ in $\mathcal{T}$}
    \State $z_t^{in} \gets [z_t; z_t]$ \Comment{Duplicate latent for CFG}
    \State $z_{in} \gets \text{concat}(z_t^{in}, [z_c; z_c])$ \Comment{Concatenate with spatial condition}
    \State Form global conditioning $C_{global}^{cond}, C_{global}^{uncond}$
    \State $\epsilon_{pred} \gets \epsilon_\theta(z_{in}, t, [C_{global}^{uncond}; C_{global}^{cond}], [c_{seq}^{uncond}; c_{seq}^{cond}])$
    \State $\epsilon_{uncond}, \epsilon_{cond} \gets \text{split}(\epsilon_{pred})$
    \State $\epsilon_{final} \gets \epsilon_{uncond} + \omega \cdot (\epsilon_{cond} - \epsilon_{uncond})$ \Comment{Apply guidance}
    \State $z_{t-1} \gets \mathcal{S}.\text{step}(\epsilon_{final}, t, z_t)$ \Comment{Scheduler denoising step}
\EndFor
\State $x_{out} \gets \mathcal{D}_{vae}(z_0)$ \Comment{Decode final latent to image}
\State \textbf{return} $x_{out}$
\end{algorithmic}
\end{algorithm}

\begin{algorithm}[ht]
\caption{MMFace-DiT Training with Rectified Flow Matching Objective}
\label{alg:rfm_training}
\begin{algorithmic}[1]
\State \textbf{Input:} DiT model $v_\theta$, VAE Encoder $\mathcal{E}_{vae}$, CLIP Text Encoder $\mathcal{E}_{text}$
\State \textbf{Require:} Dataset $D$ of triplets $(x, c_{sp}, p)$ and modality flag $m$
\Repeat
    \State Sample a batch $(x, c_{sp}, p, m)$ from $D$
    \State $z_1 \gets \mathcal{E}_{vae}(x)$; \quad $z_c \gets \mathcal{E}_{vae}(c_{sp})$ \Comment{Encode images to data latents}
    \State $c_{pooled}, c_{seq} \gets \mathcal{E}_{text}(p)$ \Comment{Encode text prompt}
    \State Sample noise latent $z_0 \sim \mathcal{N}(0, \mathbf{I})$
    \State $v_{target} \gets z_1 - z_0$ \Comment{Define the target velocity vector}
    \State Sample time $t \sim \mathcal{U}[0, 1]$
    \State $z_t \gets (1-t)z_0 + t z_1$ \Comment{Create interpolated latent on the flow path}
    \State $z_{in} \gets \text{concat}(z_t, z_c)$ \Comment{Concatenate interpolated latent and spatial condition}
    \State Form global conditioning $C_{global}$ from $t, c_{pooled}, m$
    \State $v_{pred} \gets v_\theta(z_{in}, t, C_{global}, c_{seq})$ \Comment{Predict velocity from all conditions}
    \State $\mathcal{L} \gets \mathbb{E}[\|v_{target} - v_{pred}\|^2]$ \Comment{Compute MSE loss}
    \State Update model parameters $\theta$ using gradient descent on $\mathcal{L}$
\Until{converged}
\end{algorithmic}
\end{algorithm}

\begin{algorithm}[ht]
\caption{MMFace-DiT Inference with Rectified Flow Matching Objective}
\label{alg:rfm_inference}
\begin{algorithmic}[1]
\State \textbf{Input:} Prompt $p$, spatial condition $c_{sp}$, modality flag $m$, guidance scale $\omega$, number of steps $N$
\State \textbf{Require:} Trained DiT model $v_\theta$, VAE Encoder/Decoder $\mathcal{E}_{vae}, \mathcal{D}_{vae}$, CLIP Encoder $\mathcal{E}_{text}$
\State $z_c \gets \mathcal{E}_{vae}(c_{sp})$ \Comment{Encode spatial condition}
\State $c_{pooled}^{cond}, c_{seq}^{cond} \gets \mathcal{E}_{text}(p)$ \Comment{Encode conditional prompt}
\State $c_{pooled}^{uncond}, c_{seq}^{uncond} \gets \mathcal{E}_{text}(\text{" "})$ \Comment{Encode unconditional (null) prompt}
\State Sample initial latent $z_0 \sim \mathcal{N}(0, \mathbf{I})$
\State Set step size $\Delta t \gets 1/N$
\State Let $z \gets z_0$
\For{$i=0$ to $N-1$}
    \State $t \gets i \cdot \Delta t$ \Comment{Current time}
    \State $z^{in} \gets [z; z]$ \Comment{Duplicate latent for CFG}
    \State $z_{in} \gets \text{concat}(z^{in}, [z_c; z_c])$ \Comment{Concatenate with spatial condition}
    \State Form global conditioning $C_{global}^{cond}, C_{global}^{uncond}$
    \State $v_{pred} \gets v_\theta(z_{in}, t, [C_{global}^{uncond}; C_{global}^{cond}], [c_{seq}^{uncond}; c_{seq}^{cond}])$
    \State $v_{uncond}, v_{cond} \gets \text{split}(v_{pred})$
    \State $v_{final} \gets v_{uncond} + \omega \cdot (v_{cond} - v_{uncond})$ \Comment{Apply guidance to velocity}
    \State $z \gets z + v_{final} \cdot \Delta t$ \Comment{Euler method ODE step}
\EndFor
\State $x_{out} \gets \mathcal{D}_{vae}(z)$ \Comment{Decode final latent at t=1 to image}
\State \textbf{return} $x_{out}$
\end{algorithmic}
\end{algorithm}

\section{VLM-Powered Data Enrichment Pipeline}
\label{sec:supp_vlm_pipeline}

A core contribution of our work is the creation of a large-scale, high-quality, and diverse set of textual annotations for the FFHQ and CelebA-HQ datasets. The performance of modern controllable generative models is critically dependent on the quality of their training data~\citesupp{betker2023improvingx}. We identified two primary data-related bottlenecks in the field: 1) the complete lack of captions for the 70,000 high-resolution images in the FFHQ dataset, and 2) the limited semantic richness of existing captions for CelebA-HQ. 

To overcome these limitations, we designed a sophisticated, two-stage automated annotation pipeline detailed below.

\subsection{Stage 1: VLM-Based Caption Generation}
\paragraph{The Multi-Prompt Strategy.} The first stage uses the powerful \textbf{InternVL3} Vision-Language Model~\citesupp{zhu2025internvl3x} to generate a base set of captions. Instead of using a single generic prompt, we developed a systematic multi-prompt strategy. For each image, we query the VLM with ten uniquely engineered prompts, each designed to elicit a different style and focus of information. This careful prompt engineering ensures that our final dataset captures a wide range of attributes. The prompts include:
\begin{itemize}
    \item \textbf{Few-shot Descriptive Prompts:} These ask for a standard, concise description, providing examples to guide the model's output format (e.g., "A professional headshot of a woman with medium-length curly brown hair...").
    \item \textbf{Structured Demographic Templates:} These prompts explicitly ask the VLM to fill in perceived demographic details, capturing information often missing from simple descriptions (e.g., "A [photo style] of a [age group] [gender] with [racial/ethnic appearance] features...").
    \item \textbf{Keyword-focused Prompts:} These request comma-separated keywords, ideal for capturing salient objects and attributes (e.g., "woman, long blonde hair, smiling, red dress...").
    \item \textbf{Detail-Oriented Prompts:} These focus on fine-grained features like accessories or specific makeup details.
\end{itemize}
This strategy ensures that our dataset contains a rich and diverse set of textual descriptions for every single image, forming a robust foundation for the next stage.

\subsection{Stage 2: LLM-Based Post-Processing and Augmentation}
\paragraph{The Two-Stage Refinement.} The raw text generated by the VLM often contains instructional artifacts, prompt remnants, or subtle factual inconsistencies (hallucinations). To address this, we developed a rigorous two-stage post-processing pipeline.

\textbf{First}, a rule-based cleaning script is applied to every raw caption. This script performs programmatic sanitation by stripping common instructional prefixes and suffixes (e.g., "Generate a caption for the image above:", "Max 75 tokens."), removing unfilled template placeholders (e.g., "[clothing/accessories]"), and standardizing punctuation and capitalization.

\textbf{Second}, we leverage a powerful Large Language Model, \textbf{Qwen3}~\citesupp{yang2025qwen3x}, to conduct the final, intelligent post-processing. This LLM stage serves a crucial dual role:
\begin{enumerate}
    \item \textbf{Refinement and Hallucination Mitigation:} The LLM is tasked with reviewing and rephrasing the cleaned VLM captions to improve their grammatical structure and coherence. By using the full set of cleaned captions for an image as context, the LLM can identify and correct subtle factual inconsistencies that a rule-based script would miss, thereby mitigating hallucinations.
    \item \textbf{Generation and Gap-Filling:} In cases where the initial VLM prompting and rule-based cleaning did not result in ten unique, high-quality captions, the LLM is prompted to generate novel captions. It is strictly constrained to use only the factual information present in the existing valid captions for that image, ensuring that the new captions are factually consistent while increasing the diversity of the phrasing.
\end{enumerate}
Finally, this entire pipeline strictly enforces that every caption is below the 77-token limit compatible with CLIP's context length. This process results in a dense, high-fidelity, and diverse textual annotation layer. The final dataset, containing 10 high-quality captions for each of the 100,000 images from FFHQ and CelebA-HQ (totaling 1 million captions), is being released publicly to benefit the research community.

\section{Qualitative Visualizations}
\begin{figure*}[ht!]
    \centering
    \includegraphics[width=\textwidth]{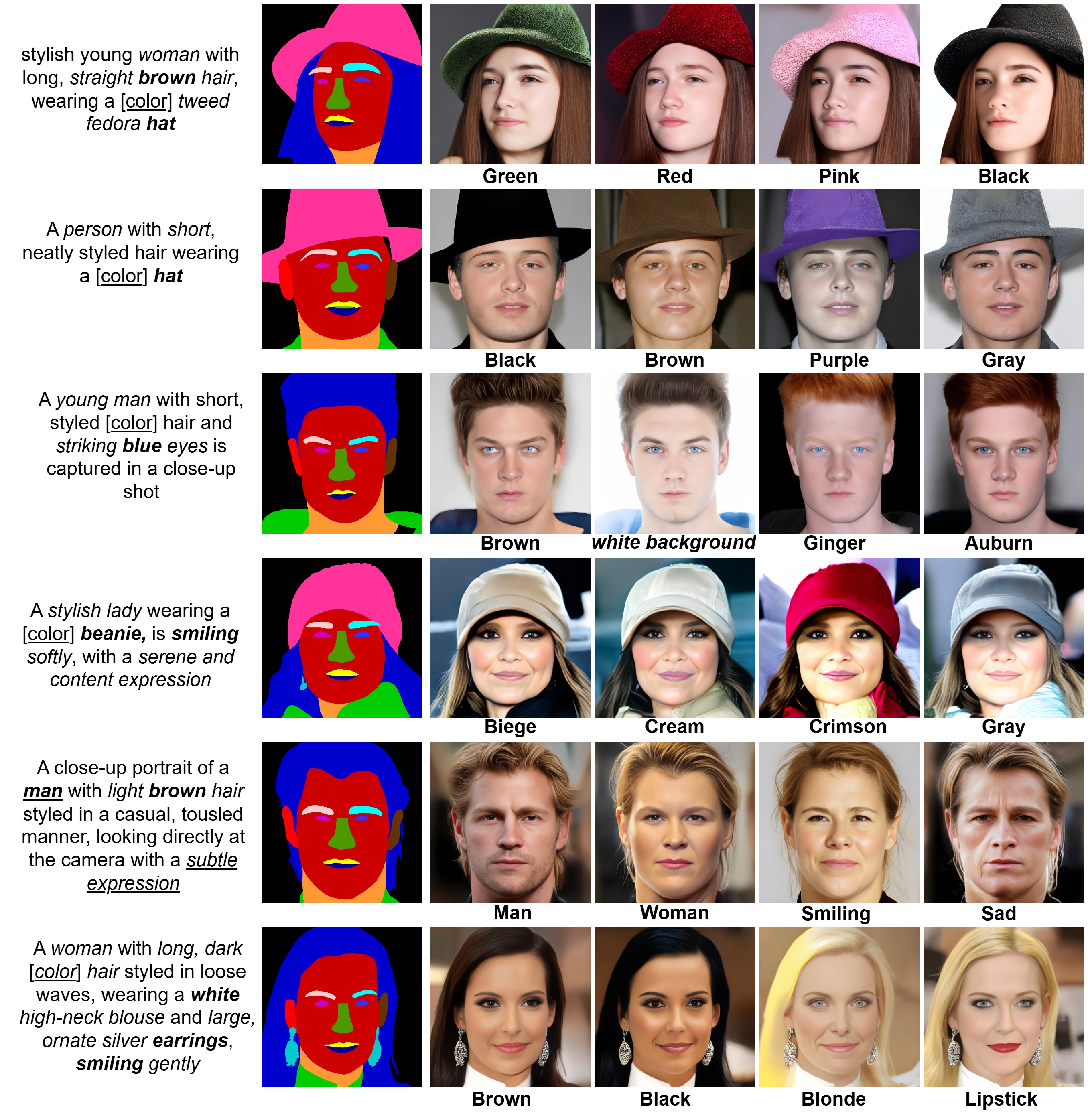}
    \caption{\textbf{Demonstration of Disentangled Fine-Grained Attribute Control.} Our MMFace-DiT exhibits exceptional disentangled control over the synthesis process. Each row is generated from a single, fixed segmentation mask, where we systematically vary a single keyword in the text prompt. The model accurately synthesizes diverse attributes—including color (hats, hair), expression (smiling, sad), gender, and even semantic concepts like background details, showcasing our model's advanced capability for precise, text-guided semantic generation.}
    \label{fig:supp_attribute_control}
\end{figure*}

\begin{figure*}[ht!]
    \centering
    \includegraphics[width=\textwidth]{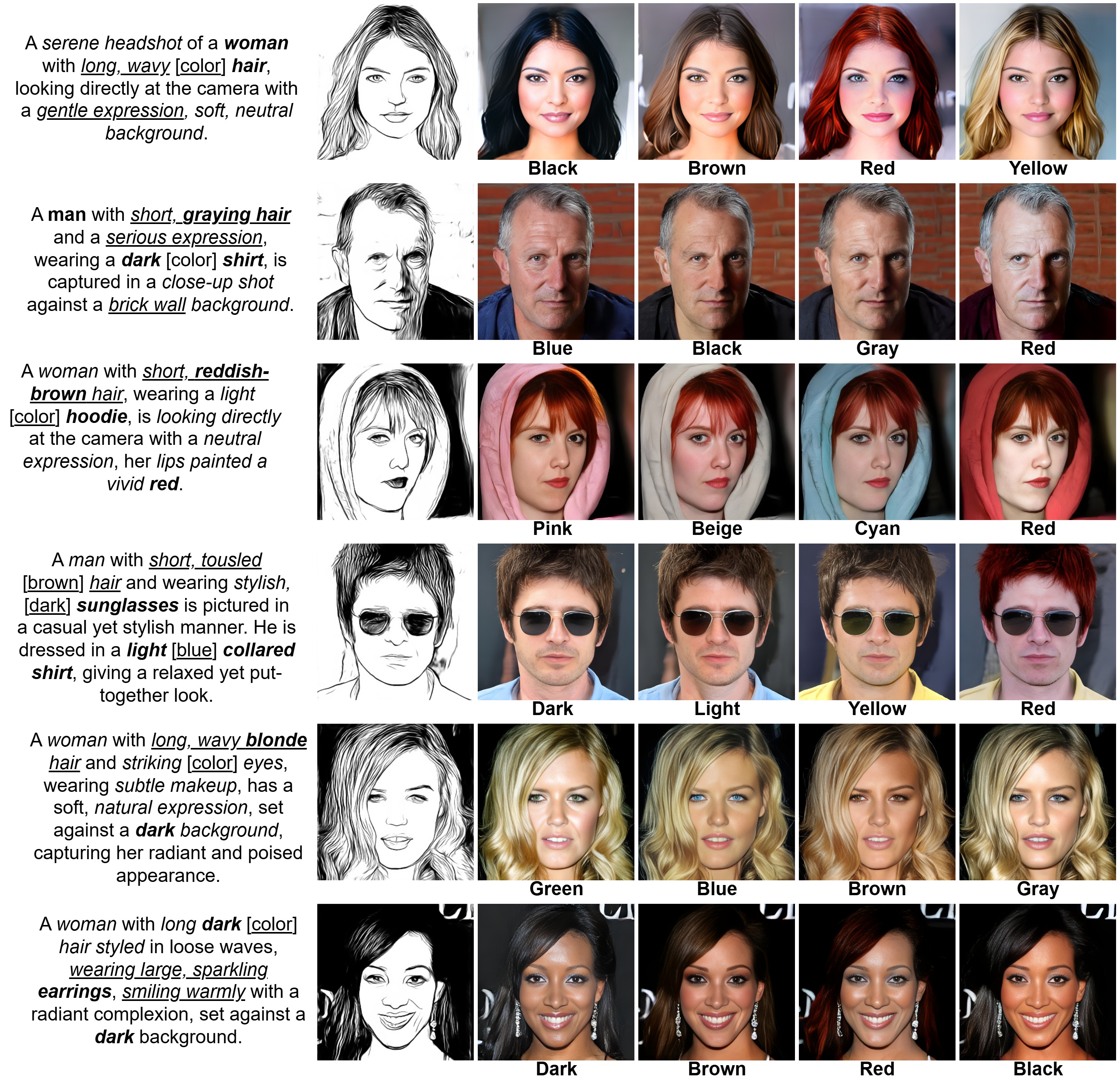}
    \caption{\textbf{Disentangled Attribute Control via Sketch-Conditioned Generation.} MMFace-DiT exhibits fine-grained disentanglement in multimodal synthesis when guided by sketch-based spatial priors. Each row is generated from a single fixed sketch while systematically varying a single textual attribute (e.g., hair color, shirt color, eye color). The model precisely follows the specified text-based edits while preserving identity, expression, and geometric consistency dictated by the sketch. This demonstrates MMFace-DiT’s capability for precise semantic integration with strong geometric priors.}
    \label{fig:supp_sketch_attribute_control}
\end{figure*}

\begin{figure*}[ht!]
    \centering
    \includegraphics[width=\textwidth]{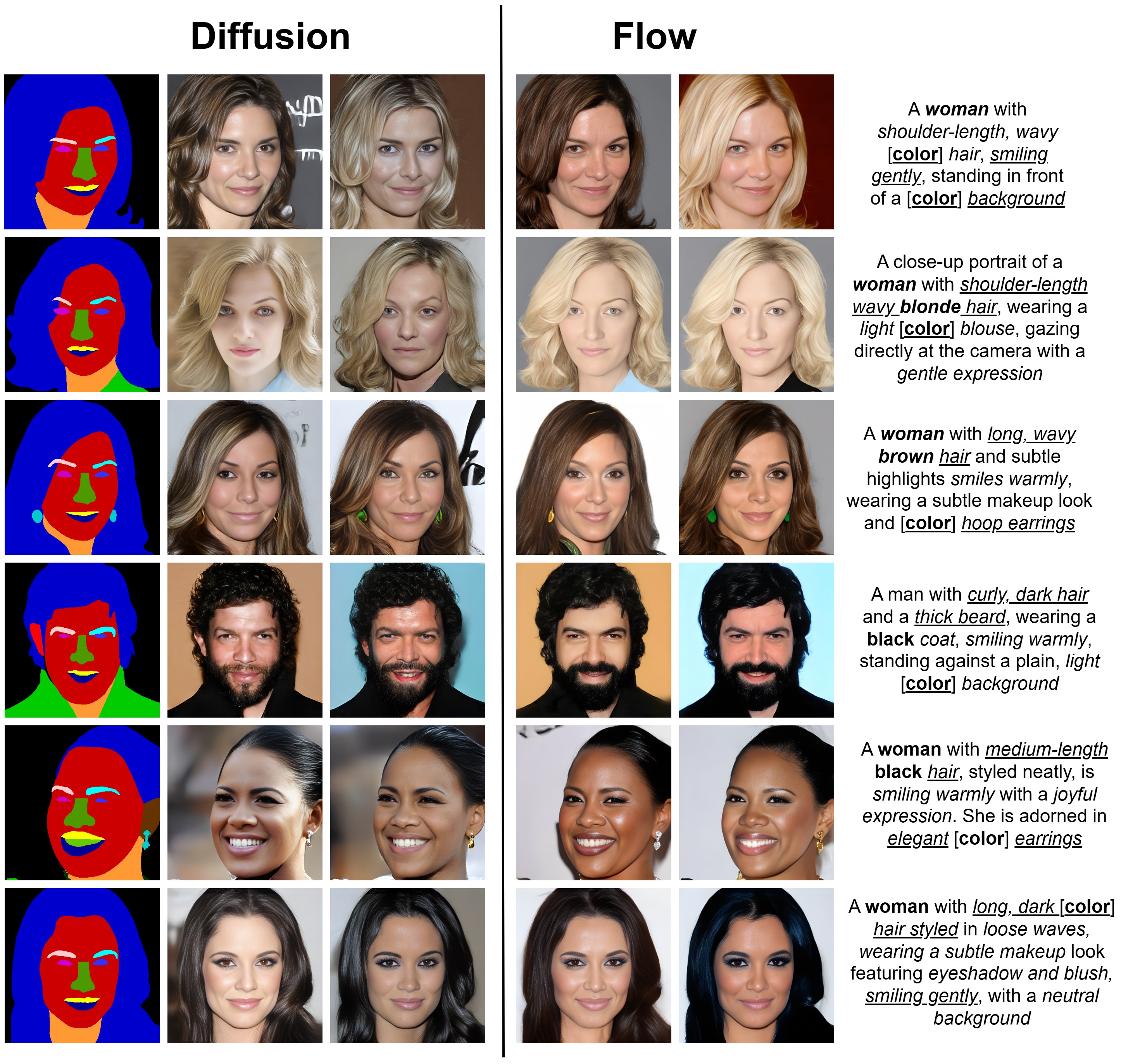}
    \caption{\textbf{Mask-Conditioned Synthesis with Diffusion and Flow Paradigms.} This figure showcases the high-quality performance of our MMFace-DiT model when trained under both Diffusion and Rectified Flow (Flow) objectives. Each example is generated from an identical segmentation mask (far left) and text prompt (far right), demonstrating the model's robust ability to synthesize diverse and realistic portraits that align with both spatial and semantic guidance. Both training paradigms yield excellent results, successfully interpreting complex attributes like hair style, expression, and accessories. Notably, the Flow-based model often exhibits a particularly refined level of photorealism, producing images with remarkably consistent lighting, skin texture, and fine-grained detail.}
    \label{fig:mask_comparison}
\end{figure*}

\begin{figure*}[ht!]
    \centering
    \includegraphics[width=\textwidth]{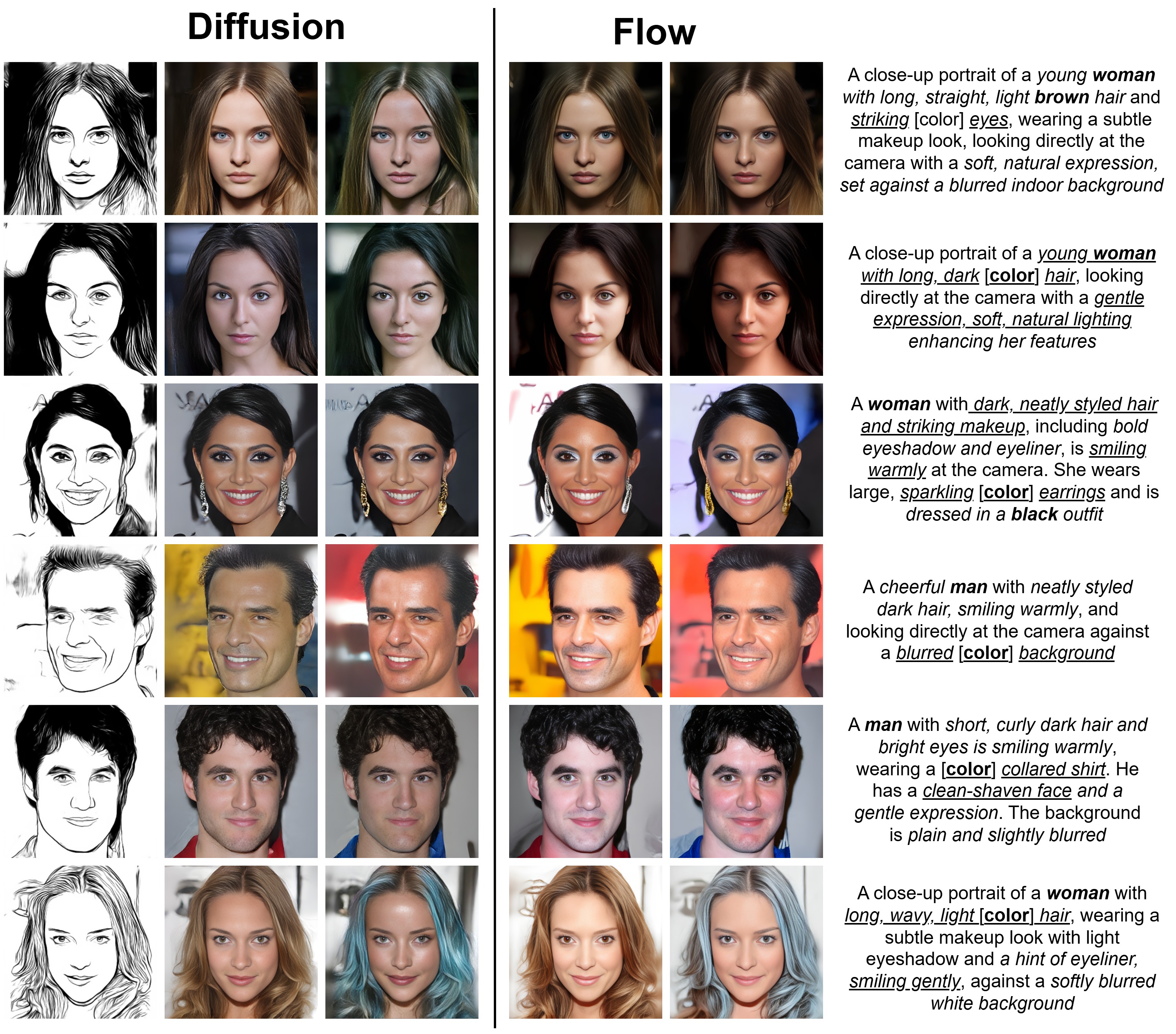}
    \caption{\textbf{Sketch-Conditioned Synthesis with Diffusion and Flow Paradigms.} This figure highlights the versatility of the MMFace-DiT architecture in translating artistic sketches into photorealistic faces, comparing results from both the Diffusion and Flow training frameworks. Conditioned on the same input sketch (far left) and text prompt (far right), both models successfully preserve the core identity, pose, and expression of the sketch while integrating the specified textual attributes. This demonstrates the model's strong multimodal capabilities regardless of the training objective. The Flow-based generations, in particular, show a strong proficiency in maintaining structural fidelity to the sketch, resulting in outputs that seamlessly blend the artistic input with photorealistic rendering.}
    \label{fig:sketch_comparison}
\end{figure*}

\begin{figure*}[h!]
    \centering
    \includegraphics[width=\textwidth]{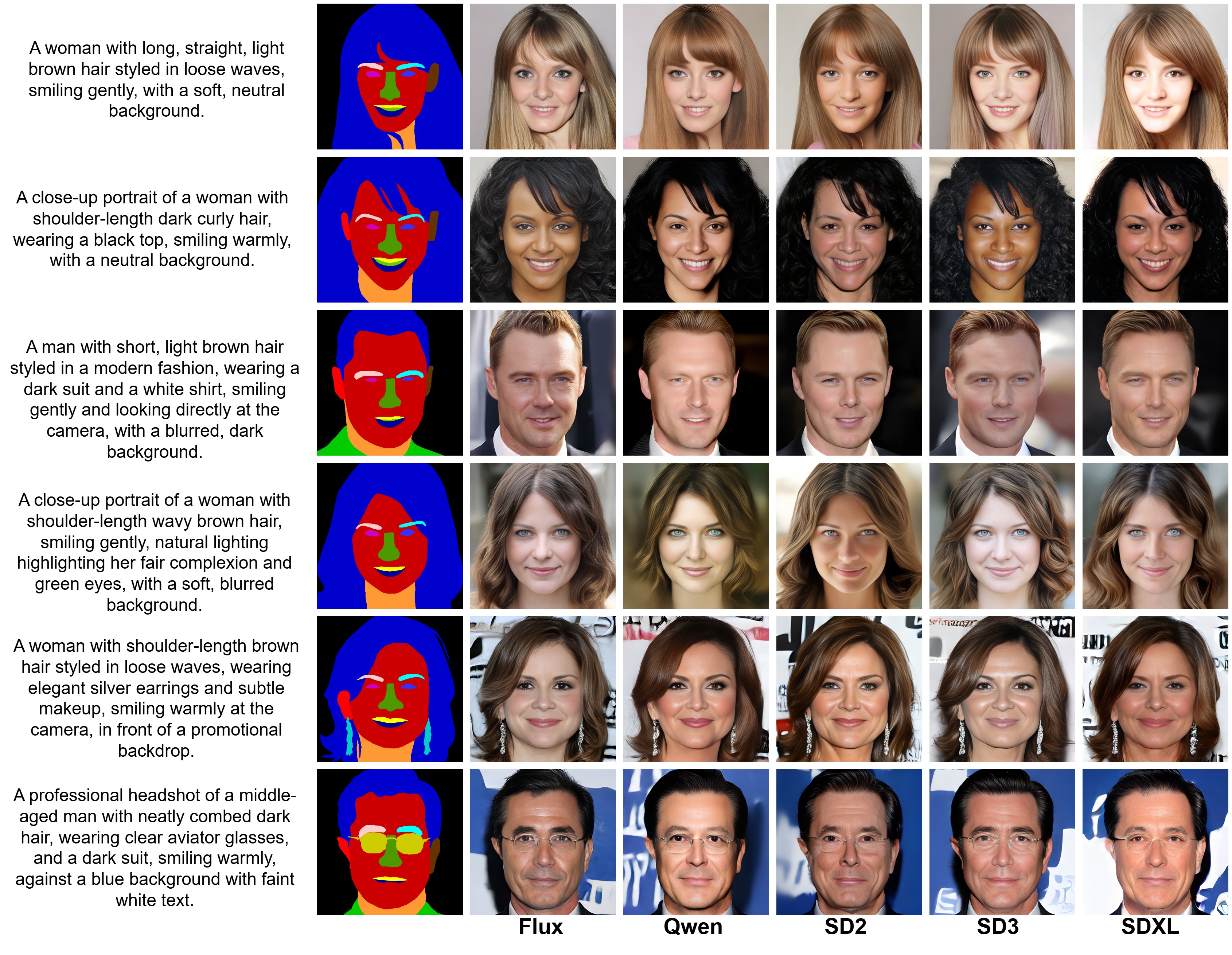}
    \caption{\textbf{Qualitative VAE Ablation with Mask-Conditioning.} This figure provides a visual comparison of five different VAE backbones integrated into our MMFace-DiT framework, using segmentation masks as spatial guidance. For each row, every model receives the identical text prompt and mask. The results illustrate a clear trade-off between statistical fidelity and perceptual realism. While models like \textbf{SD3} produce sharp outputs, they often introduce an artificial glossiness and oversaturate skin tones. \textbf{SD2} and \textbf{SDXL} can lead to desaturated or less vibrant results. In contrast, the \textbf{Flux} VAE consistently delivers the most balanced and photorealistic portraits, excelling in color accuracy, natural skin texture, and fine-detail preservation. These qualitative findings strongly establish \textbf{Flux} as the superior backbone for generating high-fidelity, artifact-free portraits.}
    \label{fig:vae_ablation_masks}
\end{figure*}

\begin{figure*}[h!]
    \centering
    \includegraphics[width=\textwidth]{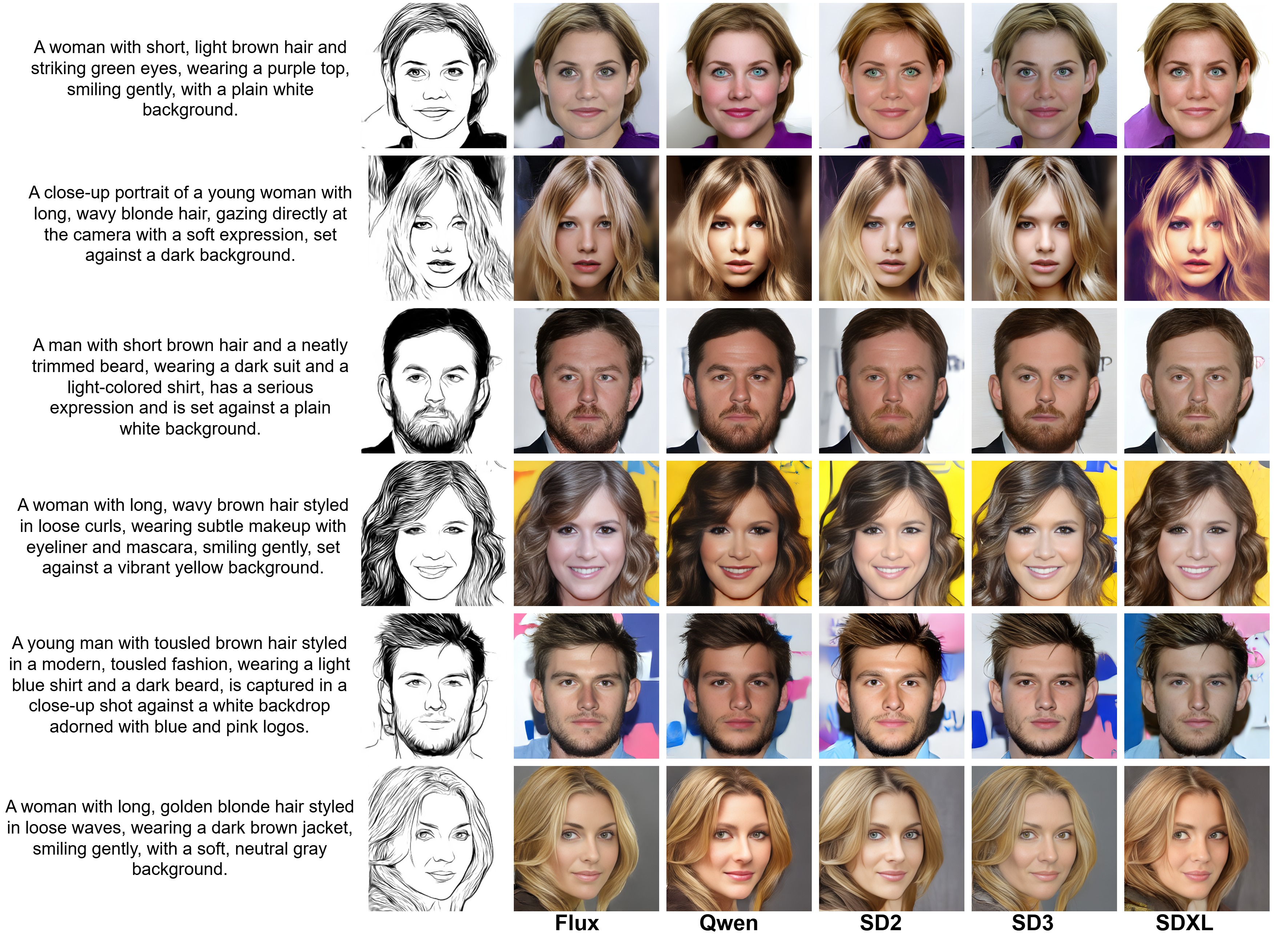}
    \caption{\textbf{Qualitative VAE Ablation with Sketch-Conditioning.} This figure showcases the performance of the five VAE backbones when conditioned on artistic sketches. Each model generates an image from the same input sketch and text prompt, testing its ability to preserve geometric structure while synthesizing photorealistic detail. The comparison highlights key differences in semantic adherence and realism. For example, several models fail to accurately render specified features like the \textit{striking green eyes} in the first row. Although \textbf{SD3} captures details sharply, its outputs can appear airbrushed. The \textbf{Flux} model demonstrates a superior synthesis capability, faithfully translating the sketch's identity and expression while accurately integrating nuanced textual details. This visual evidence aligns with our quantitative results, where \textbf{Flux} achieved the best perceptual quality (LPIPS) for sketch-conditioned generation.}
    \label{fig:vae_ablation_sketches}
\end{figure*}

\begin{figure*}[h!]
    \centering
    \includegraphics[width=\textwidth]{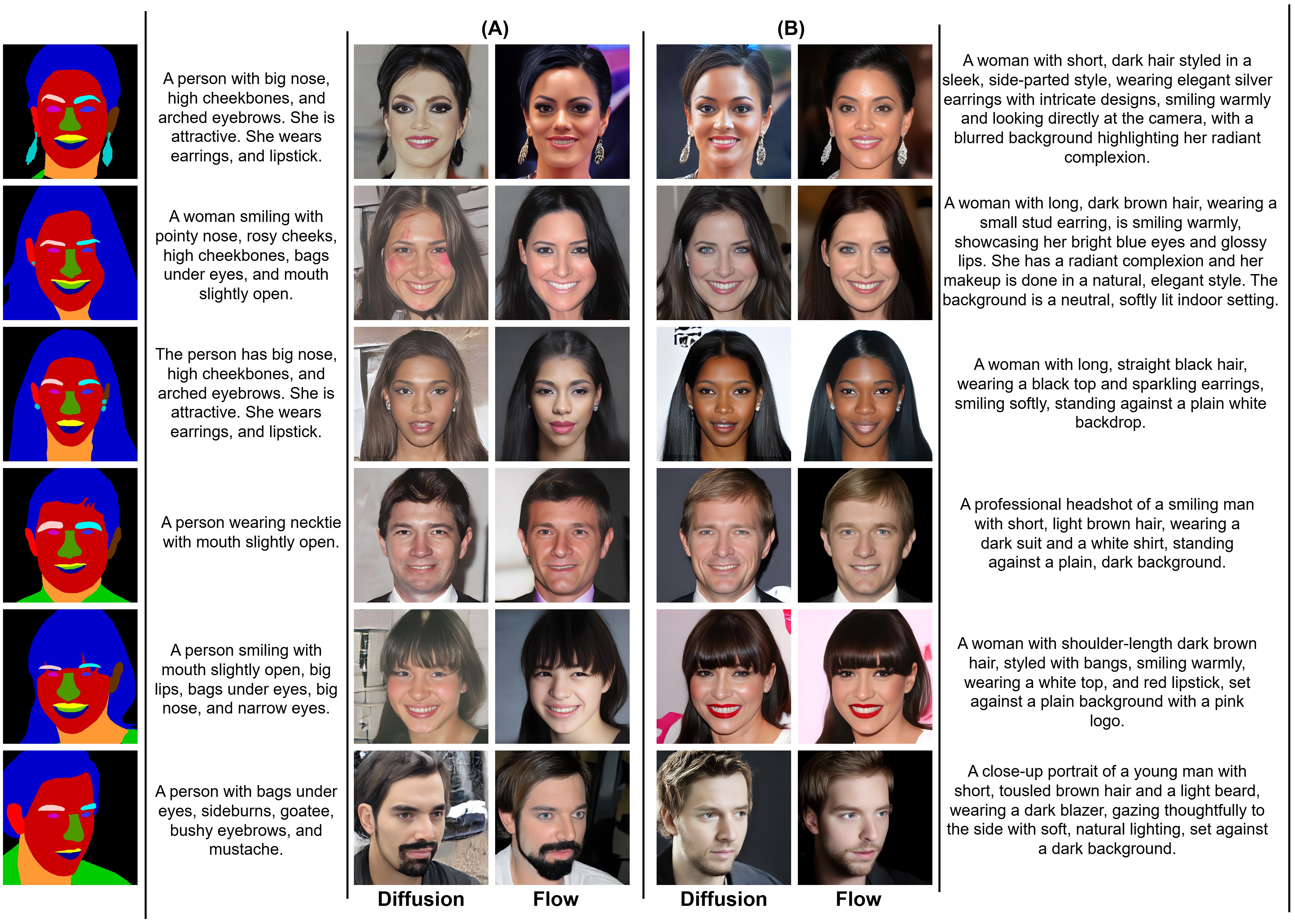}
    \caption{\textbf{Efficacy of Rich Textual Conditioning in Mask-Guided Synthesis.} This figure illustrates the substantial qualitative gain achieved through our VLM-powered data enrichment. Column \textbf{(A)} relies on the original, sparse annotations, which frequently result in flat lighting or visual artifacts (e.g., the unnatural skin texture in row 2). Conversely, Column \textbf{(B)} utilizes our comprehensive descriptions on the exact same segmentation masks. The enriched prompts empower the model to generate intricate accessories and textures that were previously absent, such as \textit{elegant silver earrings} (row 1) or a \textit{dark suit} (row 4). Furthermore, specific stylistic attributes like \textit{red lipstick} (row 5) and environmental context like a \textit{softly lit indoor setting} (row 2) are rendered with high fidelity, demonstrating that detailed semantic guidance is essential for resolving ambiguity in mask-to-image generation.}
    \label{fig:prompt_comparison} 
\end{figure*}

\begin{figure*}[h!]
    \centering
    \includegraphics[width=\textwidth]{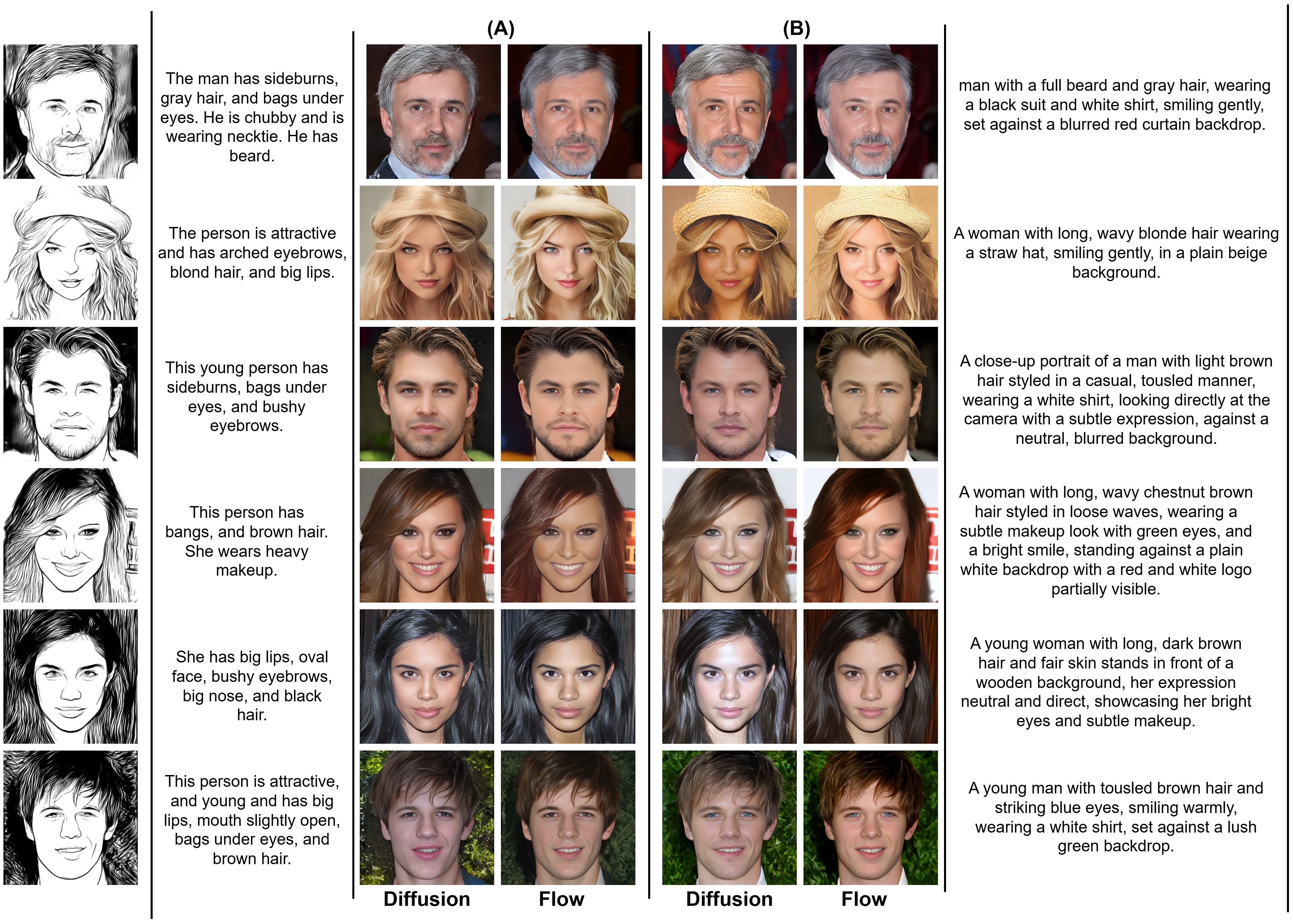} 
    \caption{\textbf{Efficacy of Rich Textual Conditioning in Sketch-Based Synthesis.} This figure evaluates the impact of our VLM-enriched annotations on sketch-to-image generation. Column \textbf{(A)} presents results using the original, brief captions, which typically yield generic attributes and unspecified settings. In contrast, Column \textbf{(B)} utilizes our comprehensive descriptions with the exact same input sketches. The enriched prompts enable the model to render precise semantic details—ranging from specific environmental contexts like a \textit{plain beige background} (row 2) or \textit{lush green backdrop} (row 6), to fine-grained facial features such as \textit{striking blue eyes} (row 6) and material textures like a \textit{straw hat} (row 2). Crucially, this semantic enrichment significantly improves photorealism and scene composition while strictly adhering to the structural constraints provided by the input sketch.}
    \label{fig:prompt_comparison_sketches}
\end{figure*}

We provide extensive qualitative results to complement our quantitative analysis. \cref{fig:supp_attribute_control} and \cref{fig:supp_sketch_attribute_control}  highlight the model's capacity for fine-grained, disentangled attribute control by varying a single word in the text prompt while keeping the spatial condition fixed. \cref{fig:mask_comparison} and \cref{fig:sketch_comparison} compare the high-quality outputs from the Diffusion and Flow training paradigms, demonstrating the model's effectiveness with both objectives. Additionally, \cref{fig:vae_ablation_masks} and \cref{fig:vae_ablation_sketches} provide the visual evidence for our VAE ablation study, illustrating the superior perceptual quality and artifact reduction achieved with the Flux VAE compared to other backbones. Finally, \cref{fig:prompt_comparison} and \cref{fig:prompt_comparison_sketches} demonstrate the substantial impact of our VLM-powered data enrichment, illustrating that our semantically rich captions yield superior photorealism and detail compared to the original sparse annotations.

{
    \small
    \setcounter{NAT@ctr}{0}
    \bibliographystylesupp{ieeenat_fullname}
    \bibliographysupp{main2}
}